\documentclass[11pt, a4paper, onecolumn, copyright, goog]{handshaketemplate}

\usepackage[
  backend=biber,
  style=authoryear,
  natbib=true,
  maxcitenames=2,
  maxbibnames=99,
  uniquelist=false,
  uniquename=false,
  dashed=false,
  mincrossrefs=99,
]{biblatex}
\DeclareDelimFormat{nameyeardelim}{\addcomma\space}
\DeclareDelimFormat{multinamedelim}{\addsemicolon\space}
\DeclareDelimFormat{finalnamedelim}{\addspace\&\space}
\DeclareDelimFormat{multicitedelim}{\addsemicolon\space}

\AtEveryBibitem{\clearfield{urldate}\clearfield{urlyear}\clearfield{urlmonth}\clearfield{urlday}}

\newif\ifvariantneurips
\newif\ifvarianticlr
\newif\ifvarianthandshake
\newif\ifvariantworkshop

 \varianthandshaketrue

\usepackage{graphicx}   %
\usepackage{multirow}   %
\usepackage{rotating}   %
\usepackage{footnote}   %
\makesavenoteenv{tabular}
\makesavenoteenv{table}

\usepackage{csquotes}

\usepackage{tikz}
\usetikzlibrary{calc}
 
\definecolor{darkblue}{rgb}{0, 0, 0.5}
\hypersetup{colorlinks=true, citecolor=darkblue, linkcolor=darkblue, urlcolor=darkblue}

\newcommand{\cornerlogo}[2][1]{%
  \begin{tikzpicture}[remember picture,overlay]
    \node[anchor=north west, xshift=2.0cm, yshift=-1.8cm] at (current page.north west)
      {\includegraphics[scale=#1]{#2}};
    \draw[line width=0.3pt]
      ($(current page.north west)+(2.0cm,-2.7cm)$) -- ($(current page.north east)+(-2.0cm,-2.7cm)$);
  \end{tikzpicture}%
}

\title{A Cheap Verifier is Good Enough: LLM Post-training is Robust to Erroneous Rewards}

\renewcommand{\today}{}

\makeatletter
\renewcommand\AB@authnote[1]{\textsuperscript{#1}}
\renewcommand\AB@affilnote[1]{\textsuperscript{#1}}
\makeatother

\author[1,2]{Andreas Plesner}
\author[2]{Curtis Northcutt}
\author[2]{Francisco Guzm\'{a}n}
\author[2]{Anish Athalye}
\affil[1]{ETH Zurich}
\affil[2]{Handshake AI}

\begin{abstract}
  When post-training large language models on tasks with semi-verifiable rewards, there are many factors (training steps, base model size, training order, data quality, verifier accuracy, etc.) that practitioners must contend with to maximize model performance. Yet, it remains unclear how well verifier agreement predicts post-training performance on such tasks. In this paper, we explore this question with over 11k H100 GPU-hours, across HealthBench and PRBench tasks in medical, legal, and finance domains.
  Across the tested domains, Qwen3 trainees (1.7B--8B on HealthBench; 8B on PRBench), evaluation splits, and frontier LLM reference judges (which we call golden verifiers), higher verifier agreement does not consistently identify the best training verifier. Expensive verifiers need not outperform inexpensive ones, and open-weight Gemma verifiers produce strong training outcomes. We compare two low-cost choices retrospectively --- a \emph{cost-reducing} choice and a \emph{balanced} choice --- with estimated grading cost reductions of $98.8\%$--$99.7\%$ relative to the golden grading protocols and average post-training score gaps of $1$--$3$ points from the best evaluated training verifier. These averages include larger losses in individual settings; they do not establish that verifier choices are interchangeable.
 \end{abstract}

\begin{document}

\cornerlogo[0.27]{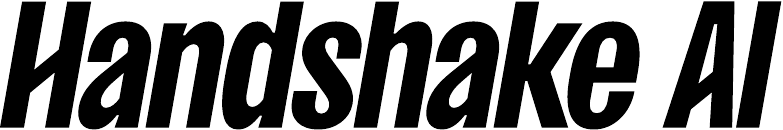}

\maketitle

\section{Introduction}

Reinforcement Learning with Verifiable Rewards (RLVR) has become the post-training method of choice for eliciting reasoning capabilities in Large Language Models~\citep{shaoDeepSeekMathPushingLimits2024,deepseek-aiDeepSeekR1IncentivizingReasoning2025,mai_thinking_1, kimiteamKimiK3Open2026, DeepSeekV41Flash2026}. 
Early work focused on domains such as mathematics and coding, where deterministic checks provide a near-perfect reward signal~\citep{austinProgramSynthesisLarge2021}. 
The field has since expanded from verifiable rewards to semi-verifiable rubric-based rewards for post-training. This extension allows training objectives to incorporate preferences and answer style alongside correctness.
A verifier (sometimes called a grader or judge) is used to score task outputs according to a rubric composed of criteria~\citep{gunjalRubricsRewardsReinforcement2025,heAdvancedIFRubricBasedBenchmarking2025,viswanathanChecklistsAreBetter2025,zhouBreakingExplorationBottleneck2025,lauBankerToolBenchEvaluatingAI2026}.
In some cases, these verifiers are used directly to provide rewards for post-training, where a natural hypothesis is that the accuracy of the rewards given by the verifier has a significant impact on the final performance of the post-trained model, especially in challenging domains with semi-verifiable rewards where taste preferences come into play. %

This hypothesis has a natural human corollary: if Student A has a TA who grades their homework $30\%$ incorrectly and Student B has a ``better'' TA who grades their homework $10\%$ incorrectly, the common hypothesis is that Student B is likely to learn better over the course of the semester. 
Across the models we post-train (students), datasets (course material), verifiers (TAs), and golden verifiers (imperfect answer keys), better agreement with the answer key does not always identify the TA whose students learn most. Agreement remains informative, particularly for identifying very weak verifiers, but does not fully determine training outcomes. Our practical question is whether inexpensive verifiers can deliver strong post-training performance within a candidate pool that also includes strong, expensive models.

While prior work has explored post-training robustness to synthetically introduced error in verifiable rewards~\citep{plesnerImperfectVerifierGood2026,caiReinforcementLearningVerifiable2025}, we examine the semi-verifiable domain where imperfect verifiers are used in practice for rubrics as rewards.

This paper evaluates verifier selection for post-training on semi-verifiable tasks.
We compare across model size, dataset domain, evaluation hardness (hard and a mixture of hard and non-hard tasks), reference choice (golden judges), and several verifiers.

\textbf{Contributions}: Across the tested combinations of these settings, we post-train an LLM and measure the performance impact.
Across the tested settings, inexpensive verifiers can deliver scores close to the best evaluated training verifier, including in comparisons with strong, expensive candidates such as Grok 4.20. Higher verifier agreement does not always identify the best choice, although very weak verifiers can harm training.
For example, the open-weight Gemma 4 31B and 26B verifiers produce some of the strongest training outcomes (\Cref{tab:js-overall,sec:healthbench tables}).
We compare two low-cost verifier choices retrospectively, with estimated grading cost reductions of $98.8\%$--$99.7\%$ relative to the golden grading protocols (\Cref{sec:practical-implications}). These reductions would translate to potentially tens of thousands of dollars per training run just for our smaller scale experiments.

\section{Methodology}
Reinforcement learning (RL), including RL from human feedback (RLHF), RL with verifiable rewards (RLVR), and RL with rubrics as rewards (RLRR) for post-training language models, works by giving rewards based on generated responses.
One of the central difficulties in RL lies in shaping these rewards: what you reward and how much.
RLVR often focuses on mathematics and coding---domains where answers can be verified deterministically, for example by checking against a known solution.
This simplifies outcome verification, but does not by itself determine what to reward or by how much. Besides correctness, one might want to reward style choices, consistency, or reasoning or answer lengths \citep{mai_thinking_1,mimo_v2_6_2026}.
To extend RL post-training to open-ended tasks where correctness is not binary, recent work has turned to \emph{rubric-based} rewards~\citep{gunjalRubricsRewardsReinforcement2025, viswanathanChecklistsAreBetter2025}.
Rubrics decompose a response's quality into explicit, interpretable criteria, each of which can be assessed by an LLM judge and combined into a scalar reward.
This structures the reward signal in a way that generalizes beyond verifiable domains, at the cost of introducing a strong dependency on the quality of the verifier.

\subsection{RL with semi-verifiable rewards}
RL with semi-verifiable rewards replaces deterministic checks, such as unit tests for code, with a judge's assessment of rubric criteria.
Formally, given an input prompt $x_i\in\mathcal{D}$, we generate a response $\hat{y}_i \sim \pi_{\theta}(\cdot \mid x_i)$ from a policy $\pi_{\theta}$.
Each prompt is associated with a rubric $R_i = \{(c_{ij}, w_{ij})\}_{j=1}^{k_i}$, a set of $k_i$ criteria $c_{ij}$ with signed weights $w_{ij} \in \mathbb{R}$.\footnote{Positive weights reward responses that satisfy the criterion; negative weights penalize responses that exhibit the behavior described by the criterion.} The role of a verifier $V_{\phi}$ is to assess whether a response satisfies each criterion,
\begin{equation}
    V_{\phi}(x_i,\,\hat{y}_i,\, c_{ij}) \;\in\; \{0, 1\},
    \label{eq:verifier}
\end{equation}
where $1$ indicates that the criterion is met and $0$ that it is not.
Following the HealthBench scoring convention~\citep{aroraHealthBenchEvaluatingLarge2025a}, these verdicts are combined into an aggregate reward via a weighted sum,
\begin{equation}
    r(x_i, \hat{y}_i) \;=\; \frac{\sum_{j=1}^{k_i} w_{ij}\, V_{\phi}(x_i,\hat{y}_i, c_{ij})}{\sum_{j:\, w_{ij} > 0} w_{ij}} \;\leq\; 1,
    \label{eq:reward}
\end{equation}
where the denominator normalizes by the total positive weight, assumed nonzero, so that a response satisfying all positive criteria and none of the negative ones achieves the maximum reward of $1$. PRBench uses a normalized variant seen in \Cref{sec: prbench scoring math}.
The RL objective is then to maximize the expected reward over the training distribution,
\begin{equation}
    \max_{\theta}\; \mathbb{E}_{x_i \sim \mathcal{D},\; \hat{y}_i \sim \pi_{\theta}(\cdot \mid x_i)}\!\left[\, r(x_i, \hat{y}_i) \,\right].
    \label{eq:rl_objective}
\end{equation}
As \Cref{eq:reward,eq:rl_objective} make clear, verifier errors influence the training reward signal. We study whether agreement with a golden verifier predicts downstream policy performance; we do not characterize which errors cause particular training outcomes.

\subsection{The verifier hierarchy}

\begin{figure}[t]
    \centering
    \includegraphics[width=\linewidth]{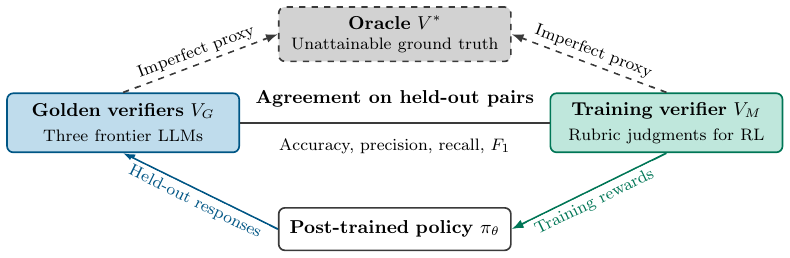}
    \caption{\textbf{Verifier hierarchy and roles.} Dashed arrows denote imperfect approximation of the unattainable oracle $V^*$. We compare training-verifier $V_M$'s judgments with golden-verifier $V_G$'s judgments on the same held-out response--criterion pairs. $V_M$ supplies training rewards; $V_G$ evaluates held-out policy responses. Agreement with a golden verifier does not establish ground-truth correctness.}
    \label{fig:verifier-hierarchy}
\end{figure}

Determining whether a response satisfies a criterion (e.g. correctness, helpfulness, safety) is itself a judgment call that no system can make with perfect fidelity. We formalize this notion through a hierarchy of verifiers, shown in \Cref{fig:verifier-hierarchy}.

\textbf{Oracle verifier.} At the top of the hierarchy sits an unattainable \emph{oracle} $V^*$, a theoretical verifier that returns ground-truth verdicts on every (response, criterion) pair.
Human annotators and LLMs approximate $V^*$ and can introduce their own noise, biases, and failure modes.
For instance, inter- and intra-annotator disagreement are the empirical signatures of human imperfection; analogous inconsistencies arise in LLM-based judges.

\textbf{Golden verifiers.} Prior work has shown that frontier LLMs can match human-to-human agreement on some evaluation tasks~\citep{zhengJudgingLLMasajudgeMTbench2023a,aroraHealthBenchEvaluatingLarge2025a}.
We use frontier models as \emph{golden verifiers} $V_G$: scalable reference judges that remain imperfect proxies for ground truth.
They let us study how training-time verifier choice affects post-training outcomes without requiring human annotation of every response.

\textbf{Training-time verifiers.} We use an LLM-based \emph{training-time verifier} $V_M$ to drive the policy gradient.
We measure agreement with the golden verifiers, following the reference-judge setup of \citet{liuExaminingReasoningLLMsasJudges2026}. Agreement is not ground-truth correctness. The golden verifiers also score the trained policies; their grading cost precluded using them for training. The training pool nevertheless includes strong, expensive models such as Grok 4.20. Using the same references for agreement and policy evaluation defines a common evaluation target, but does not rule out shared judge biases as policy outputs change during training.
Verifier alignment is quantified by accuracy, precision, recall, and $F_1$ between $V_M$ verdicts and $V_G$ verdicts on held-out (response, criterion) pairs. For each dataset, these pairs come from a fixed slice of $20$ \texttt{hard} tasks for the \texttt{hard} split, plus $20$ \texttt{non-hard} tasks for the mixed \texttt{all} split.
To reduce dependence on any particular reference, we evaluate against three frontier models from different vendors and report averages across them.\footnote{We compare against each judge's labels separately and then average the metrics, rather than combining labels first.}
We post-train each policy using its assigned training-time verifier to score rollout responses, and use the golden verifiers to score evaluation responses.

We batch multiple criteria in each verifier call and refer to the maximum number of criteria per call as the verifier's batch size, distinct from the policy-training batch size.

\subsection{Experimental setup}
\label{sec: experimental setup}
We focus our experiments on three task domains: medical, finance, and legal tasks. The medical tasks are from HealthBench~\citep{aroraHealthBenchEvaluatingLarge2025a}, while the finance and legal tasks are from PRBench~\citep{akyurekPRBenchLargeScaleExpert2025}.

\textbf{Data.} All the datasets contain subsets labeled as hard. We denote the split with the hard tasks as \texttt{hard} and the split with the remaining as \texttt{non-hard}.
We use the \texttt{hard} and 10\% of the \texttt{non-hard} tasks for evaluation, and leave the remaining non-hard problems for training (the \texttt{non-hard} are split into training and evaluation using a fixed seed of 42).
Per-domain split sizes are listed in \Cref{tab: dataset sizes}.
Some PRBench prompts are very long, so we drop the 3 tasks exceeding $90{,}000$ characters to keep prompts and responses within Qwen3's $32{,}768$-token native context (\Cref{sec: prbench filtering}); \Cref{tab: dataset sizes} reflects this filter.

\begin{table}[t]
    \centering\small
    \caption{Available dataset pools for training and evaluation. ``Hard eval-only'' contains only \texttt{hard} tasks; ``Total eval'' also includes held-out \texttt{non-hard} tasks. Policy scoring uses subsets of these pools.
    }
    \begin{tabular}{lccc}
    \toprule
    \textbf{Domain} & \textbf{Train ($n$)} & \textbf{Hard eval-only ($n$)} & \textbf{Total eval ($n$)} \\
    \midrule
    HealthBench       & $3600$ & $1000$ & $1400$ \\
    PRBench legal     & $225$ & $249$ & $274$  \\
    PRBench finance   & $269$ & $299$ & $329$  \\
    \bottomrule
    \end{tabular}
    \label{tab: dataset sizes}
\end{table}

\textbf{Trainees} are \textbf{Qwen3 1.7B}, \textbf{4B}, and \textbf{8B}~\citep{yangQwen3TechnicalReport2025}, all initialized from the Hugging Face checkpoints. We vary trainee size on HealthBench; PRBench uses Qwen3 8B.

\textbf{Training-time verifiers ($\mathbf{V_M}$).}
We tested a wide selection of pretrained models on HealthBench generations from Qwen3 8B on the \texttt{hard} subset; the full screening results for all candidates are in \Cref{app:verifier-screening}.
From the $52$ tested models, we select 12 for further experiments. These are: Gemma 4 31B, Gemma 4 26B, GPT oss 120B, GPT oss 20B, Gemini 3 flash preview, Grok 4.20, Qwen3 235B, Ministral 14B, Qwen3 8B (a self-judge for the 8B trainee), Mistral 4 small, Qwen3 Next 80B, and LFM2 24B. 
API instability and grading cost prevented some runs, including runs with Grok 4.20, from completing. Incomplete runs are excluded from the policy comparisons; candidate coverage therefore varies across domains and trainee sizes. Completed runs with poor or collapsed policies remain in the result tables.
All are run at $\textrm{batch}{=}3$, $\textrm{thinking}{=}\texttt{low}$.

\textbf{Golden verifiers ($\mathbf{V_G}$).} Our headline golden is \textbf{Claude Opus 4.6} with $\textrm{thinking}{=}\texttt{medium}$ and $\textrm{batch}{=}1$ (one criterion per call), which we used for most early experiments.
To assess robustness to the choice of $V_G$, we additionally run two alternate goldens drawn from different model families: \textbf{Gemini 3.1 Pro} and \textbf{GPT-5.5}, both with the same grading configuration.
Re-ranking the twelve $V_M$ candidates and the trained policies under these alternate golden verifiers is purely an evaluation operation and does not require additional training.
Unless stated otherwise, golden-judge results are averaged across all three references, following the approach of \citet{vergaReplacingJudgesJuries2024}. Averaging reduces dependence on any single reference, although the judges may share biases.
Alignment metrics between the three golden verifiers, including their agreement with a $2$-of-$3$ majority aggregate, are reported in \Cref{app:golden-alignment}.

To spot-check the reference labels, two human experts in rubric data reviewed 50 HealthBench criteria graded by Opus 4.6.
Both reviewers were unable to discern any errors in Opus's labels. However, this small check does not establish perfect golden-judge accuracy.

\textbf{Training algorithm and hyperparameters} can be found in \Cref{sec:training-details}. We use GRPO with within-prompt reward standardization and no effective KL penalty; verifier differences therefore affect learning through normalized advantages.

\textbf{Policy evaluation.} The golden verifiers score trained policies on larger task sets than the agreement slices, and the two are disjoint: $100$ \texttt{hard} and $40$ \texttt{non-hard} HealthBench tasks, and $53$--$55$ PRBench tasks per domain. \Cref{app:policy-eval} gives the exact task sets, how the \texttt{all} split is weighted, and how rubric verdicts become policy scores.

\section{Verifier Agreement and Post-training Performance}
\label{sec:results}

We first compare verifier agreement with post-training performance across domains and trainee sizes. Detailed per-verifier metrics and policy scores are in \Cref{sec:healthbench tables,sec:prbench results}.

\subsection{Agreement is informative but does not fully rank training verifiers}
\label{sec:domains}

\begin{figure}[t]
\centering
\includegraphics[width=\textwidth]{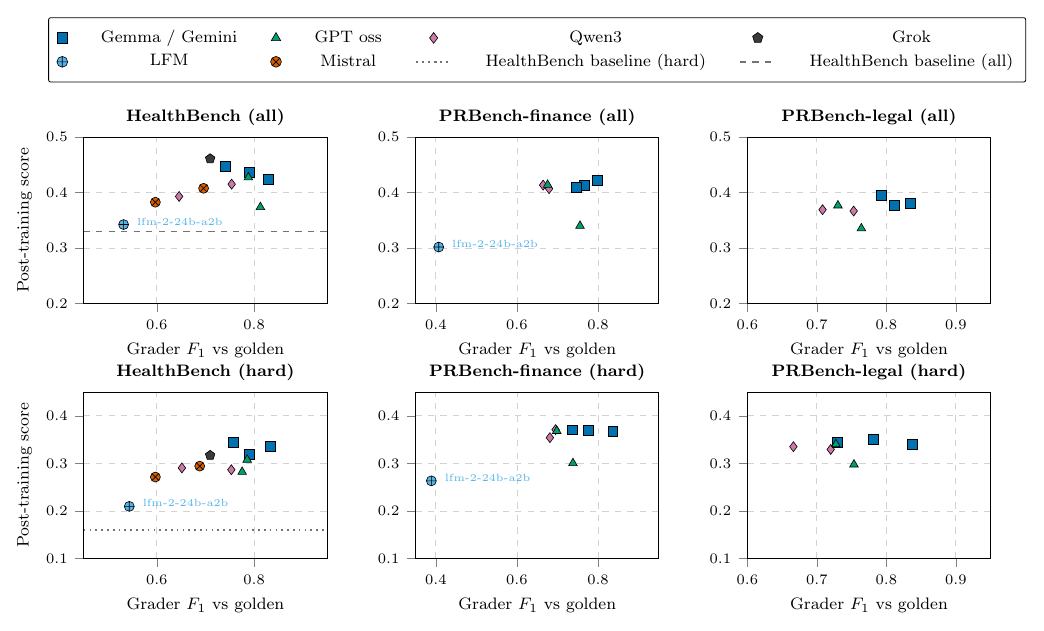}
\caption{\textbf{Higher verifier $F_1$ agreement with a golden judge does not consistently correspond to higher post-training score.}
For each dataset (HealthBench, PRBench finance, PRBench legal), every point is one (verifier, split) evaluation of a completed Qwen3 8B training run, averaged across the three golden judges. Horizontal lines mark the HealthBench no-post-training baselines. PRBench points use the clipped aggregate; normalized-score baselines are reported alongside normalized policy scores in the appendix tables (\Cref{sec: prbench scoring math,sec:prbench results}). The Qwen3 8B training verifier is omitted from this plot and \Cref{fig:score-vs-noise-trainees}; its collapsed-policy outcomes remain in the appendix tables.}
\label{fig:score-vs-f1-per-dataset}
\end{figure}

\Cref{fig:score-vs-f1-per-dataset} compares six dataset--split combinations: HealthBench, PRBench finance, and PRBench legal, each with hard and mixed evaluation sets. Agreement and policy performance can be positively associated without giving identical verifier rankings. For example, Qwen3 8B scores on HealthBench-all are $0.37$ with GPT oss 20B and $0.46$ with Grok 4.20, despite GPT oss 20B having higher $F_1$; inexpensive Gemma 4 26B scores $0.45$ (\Cref{tab: healthbench judges all}). The practical result is that low grading cost can coexist with strong training outcomes, rather than that agreement has no predictive value.

\begin{figure}[t]
\centering
\includegraphics[width=\textwidth]{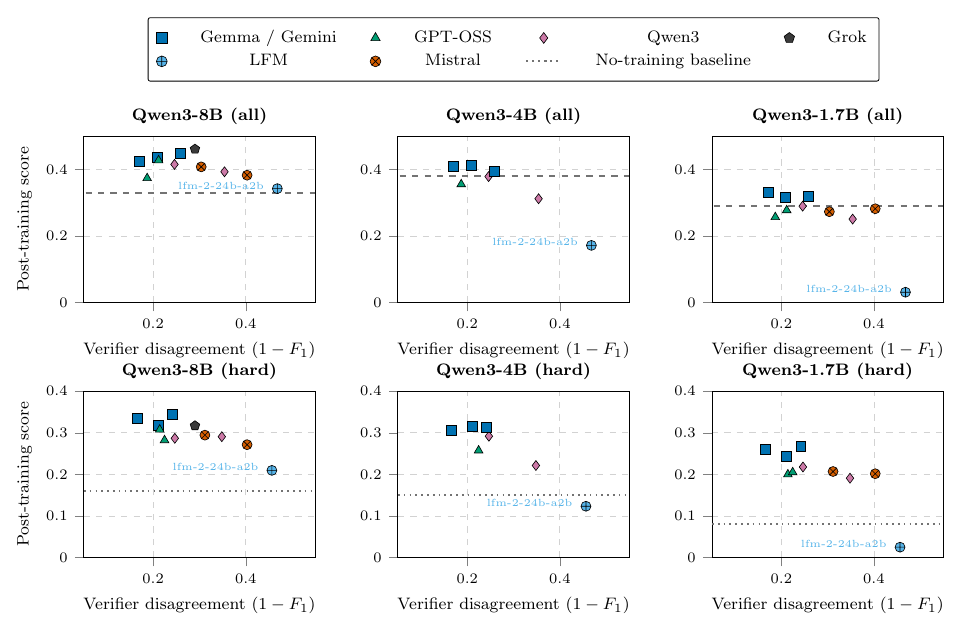}
\caption{\textbf{Verifier disagreement and post-training performance across Qwen3 trainee sizes.} Results are averaged evaluation scores across the three golden judges. The horizontal dotted line represents baseline performance for each model. Each column is a different trainee size (8B, 4B, 1.7B), with mixed and hard evaluation splits in the two rows. The $x$-axis is $1-F_1$; it does not define a noise-tolerance threshold.}
\label{fig:score-vs-noise-trainees}
\end{figure}

\Cref{fig:score-vs-noise-trainees} shows the comparison across the tested Qwen3 model sizes (up to 8B parameters).

\Cref{fig:score-vs-noise-trainees} also shows that weak verifiers can harm training: Qwen3 8B trained with LFM2 24B still clears its no-training baseline on HealthBench-hard, whereas Qwen3 1.7B trained with the same verifier falls well below baseline. Failure is not confined to low agreement: the Qwen3 8B training verifier yields below-baseline scores despite $F_1\approx0.76$ (\Cref{tab: healthbench judges hard}). These outcomes do not establish a universal quality threshold or explain the causes of failure. Here, $1-F_1$ measures disagreement with a reference, not the probability of an incorrect rubric verdict.

\subsection{Agreement, verifier capability, and training outcomes}
\label{sec:verifier-attributes}

The comparisons support a role for verifier quality without requiring a monotonic ranking by agreement. On HealthBench-hard, LFM2 24B has $F_1=0.54$ and yields low policy scores, whereas Gemma 4 31B and Gemma 4 26B have $F_1=0.83$ and $0.76$, respectively, and both yield an 8B policy score of $0.34$ (\Cref{tab: healthbench judges hard}). The same table reports GDPval as a separate measure of general model capability~\citep{patwardhanGDPvalEvaluatingAI2025}. These comparisons concern particular verifiers and do not isolate a causal effect of either capability or agreement.

\Cref{app:predictors} summarizes the descriptive value of other agreement metrics in the existing results. Accuracy and $F_1$ explain part of the observed variation in policy scores, and the strongest summary depends on the records included. Thus, evidence that agreement is informative is compatible with our cost finding: inexpensive verifiers can still be among the strongest training choices.

Criterion-level agreement also differs from the signal used for learning. The reward aggregates criteria with signed weights, while GRPO standardizes rewards within each prompt's rollout group. Unweighted $F_1$ does not measure whether those weighted rewards give the same relative advantages to competing responses. Our fixed-response screening metrics provide a practical comparison of verifiers, but do not characterize training-time errors as the policy changes. We therefore interpret the per-verifier results as observed cost--performance tradeoffs, without assigning significance to small score gaps or attributing them solely to agreement.

\section{Practical Implications for Reducing Post-training Costs}
\label{sec:practical-implications}

\begin{figure}[t]
\centering
\includegraphics[width=\textwidth]{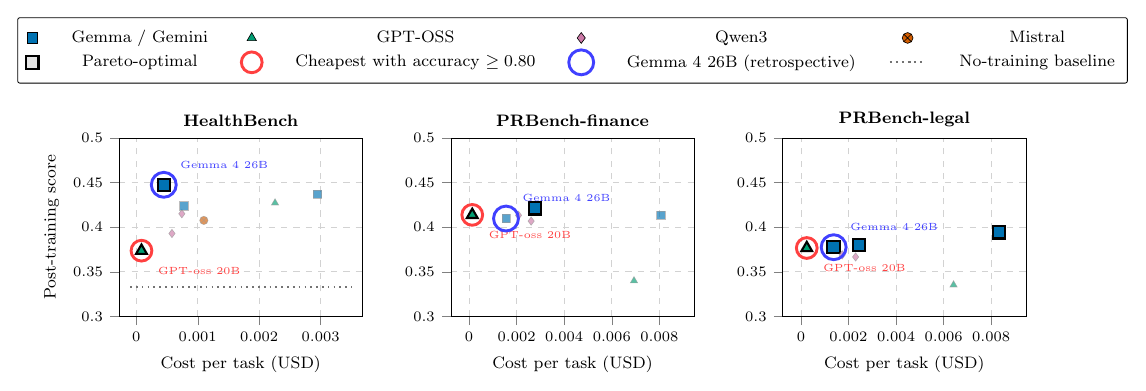}
\caption{\textbf{Cost vs.\ post-training score, per dataset, averaged across
the three goldens, with our two retrospective choices highlighted.}
One point per (dataset, verifier); panels identify datasets, and marker shape and color encode verifier provider family.
Bold-outlined markers are Pareto-optimal in (cost, post-training score) for
that dataset; smaller faded markers are sub-optimal. The dotted line marks the
HealthBench no-post-training baseline. PRBench baselines are reported on the normalized scale in the appendix tables. PRBench scores use the clipped aggregate. \textbf{Red rings}: cost-reducing Pareto verifier picks.
\textbf{Blue rings}: the fixed balanced choice, Gemma 4 26B. These examples illustrate different cost--score tradeoffs within the tested candidate pool.}
\vspace{4mm}
\label{fig:recommendations-per-dataset}
\end{figure}

The comparisons in \Cref{fig:score-vs-noise-trainees} motivate considering verifier cost alongside agreement. We compare two choices identified retrospectively: (1) the \textbf{cost-reducing Pareto verifier} --- the cheapest candidate with accuracy $\geq 0.80$ against the golden judge, falling back to the cheapest if none qualifies; and (2) the \textbf{balanced Pareto verifier} --- the fixed choice Gemma 4 26B, selected for its mean post-training score in our candidate pool. \Cref{tab:strategy-audit} compares these with other choices, and \Cref{fig:recommendations-per-dataset,fig:cost-vs-score-avg} show their cost--score tradeoff per dataset and on average. The balanced choice uses observed training outcomes; it is not an automatic selection rule validated without a post-training sweep, and neither choice guarantees Pareto optimality in a new setting.

We compare the choices using three metrics, each computed per cell (one dataset--split--golden combination), among candidates with a completed run and an available cost estimate: the \textbf{score gap}, the post-training score of the empirically best eligible verifier in the cell minus the score of the pick (lower is better; $0$ means the pick recovered the best verifier); the \textbf{uplift}, the fraction of the available post-training improvement over the no-training baseline that the pick captures; and the \textbf{cost reduction}, the fraction of per-task grading cost saved relative to the golden grading protocol. Averaged across all three datasets, both evaluation splits (\texttt{all} and \texttt{hard}), and the three golden judges, the balanced Pareto verifier achieves a $+0.010$ score gap, captures $95\%$ of the available uplift, and gives a $98.8\%$ cost reduction; the cost-reducing Pareto verifier achieves a $+0.027$ score gap, $82\%$ uplift, and a $99.7\%$ cost reduction (\Cref{tab:js-overall}). The cheaper choice has a larger score gap, including $0.062$ (6.2 points) on HealthBench-all, where it captures $46\%$ of the available uplift.

These are estimates of verifier API cost under the recorded grading protocols and pricing snapshot, not reductions in total training cost. They depend on model prices (including fallback estimates where prices were unavailable), caching, and batching; golden judges grade one criterion per call, while training verifiers grade up to three. We did not train a policy with a golden verifier, so the golden-cost comparison does not establish equivalent policy quality. \Cref{app:cost-units} reports judge calls and tokens per task.

\paragraph{Where the grading savings come from.}
\Cref{tab:grading-cost-breakdown} separates grading resource use from dollar cost on the HealthBench cost probe. The golden protocols require $2.7\times$ as many calls as the training-verifier protocols, but call reduction alone does not explain the price difference. Claude Opus 4.6 uses about $3.9\times$ as many total tokens as Gemma 4 26B while costing about $296\times$ as much on this probe. Under a matched $\textrm{batch}{=}3$, $\textrm{thinking}{=}\texttt{low}$ protocol, the token ratio falls to $1.44\times$ (\Cref{app:cost-units}). Lower token prices therefore account for most of the observed dollar savings.

\begin{table}[t]
\centering\small
\caption{\textbf{Grading resources and cost on a fixed HealthBench probe.} Calls and total input-plus-output tokens are per task; USD is per $1{,}000$ tasks at the recorded prices. Golden protocols use one criterion per call and medium thinking; training protocols use up to three criteria and low thinking. Probe sampling and pricing qualifications are in \Cref{app:cost-units}.}
\label{tab:grading-cost-breakdown}
\setlength{\tabcolsep}{7pt}
\begin{tabular}{lrrr}
\toprule
\textbf{Verifier} & \textbf{Calls/task} & \textbf{Tokens/task} & \textbf{USD/1k tasks} \\
\midrule
Claude Opus 4.6 (golden) & 11.6 & 21,321 & 133.20 \\
GPT-5.5 (golden)         & 11.6 & 17,093 & 99.70 \\
\midrule
Gemma 4 31B             & 4.3 & 5,498 & 0.77 \\
Gemma 4 26B             & 4.3 & 5,514 & 0.45 \\
GPT oss 20B             & 4.3 & 6,157 & 0.08 \\
\bottomrule
\end{tabular}
\end{table}

Self-hosting offers a complementary cost measure: Gemma 4 26B grades a HealthBench rollout in about $0.51$ H100-seconds, equivalent to roughly $14$ H100-hours for $96{,}000$ rollouts at this measured rate. These measurements make the source of the savings concrete while keeping API pricing separate from policy quality and total training cost. Actual training bills also depend on response-length growth and cache reuse. The probe supports comparison at a fixed operating point; the per-domain cost--score results remain the basis for comparing training verifiers.

\begin{table}[t]
\centering

\caption{\textbf{Verifier selection across $18$ evaluation cells}
($3$ datasets $\times$ $2$ splits $\times$ $3$ goldens; trainee Qwen3 8B).
\textbf{Splits}: \texttt{all} = \texttt{hard} $+$ \texttt{non-hard} combined with the weights in \Cref{app:policy-eval}; \texttt{hard} = the \texttt{hard} subset only.
The cost-reducing rule, fixed balanced choice, and score gap and cost reduction metrics are defined in \Cref{sec:practical-implications}.
\textbf{Uplift} $= (\text{chosen}-\text{baseline})/(\text{max}-\text{baseline})$ is the fraction of available post-training improvement captured ($100\%$ matches the best verifier per cell; $0\%$ matches no training).
\textbf{Bold} marks the better value per column between the cost-reducing and balanced picks.}
\small
\setlength{\tabcolsep}{3pt}
\renewcommand{\arraystretch}{1.20}
\resizebox{\textwidth}{!}{%
\begin{tabular}{@{}l l r r r r r r r r@{}}
\toprule
& & \multicolumn{2}{c}{\textbf{Overall}}& \multicolumn{2}{c}{\textbf{HealthBench}} &
\multicolumn{2}{c}{\textbf{PR-finance}} &
\multicolumn{2}{c}{\textbf{PR-legal}} \\ \cmidrule(r){3-4}
\cmidrule(lr){5-6}
\cmidrule(lr){7-8}
\cmidrule(l){9-10}
\textbf{Variant} & \textbf{Metric}
 & \textbf{all} & \textbf{hard}
 & \textbf{all} & \textbf{hard}
 & \textbf{all} & \textbf{hard}
 & \textbf{all} & \textbf{hard} \\
\midrule
\multirow{3}{*}{\shortstack[l]{Cheapest verifier\\(no quality floor)}}
 & Score gap      & +0.030  & +0.026  & +0.062  & +0.054  & +0.011  & +0.011  & +0.017  & +0.013  \\
 & Uplift         & 77\%    & 86\%    & 46\%    & 70\%    & 95\%    & 95\%    & 89\%    & 93\%    \\
 & Cost reduction & 99.7\%  & 99.7\%  & 99.4\%  & 99.5\%  & 99.8\%  & 99.8\%  & 99.8\%  & 99.8\%  \\
\midrule
\multirow{3}{*}{\shortstack[l]{Cost-reducing\\Pareto verifier}}
 & Score gap      & +0.030 & +0.023 & +0.062 & +0.046 & \textbf{+0.011} & \textbf{+0.011} & \textbf{+0.017} & +0.013 \\
 & Uplift         & 77\%   & 88\%   & 46\%   & 76\%   & \textbf{95\%}   & \textbf{95\%}   & \textbf{89\%}   & \textbf{93\%}   \\
 & Cost reduction & \textbf{99.7\%} & \textbf{99.7\%} & \textbf{99.4\%} & \textbf{99.1\%} & \textbf{99.8\%} & \textbf{99.8\%} & \textbf{99.8\%} & \textbf{99.8\%} \\
\midrule
\multirow{3}{*}{\shortstack[l]{Balanced\\Pareto verifier}}
 & Score gap      & \textbf{+0.012} & \textbf{+0.008} & \textbf{+0.001} & \textbf{+0.000} & +0.015  & +0.012  & +0.019  & \textbf{+0.012} \\
 & Uplift         & \textbf{93\%}   & \textbf{96\%}   & \textbf{99\%}   & \textbf{100\%}  & 94\%    & \textbf{95\%}   & 87\%    & \textbf{93\%}   \\
 & Cost reduction & 98.8\%  & 98.9\%  & 98.1\%  & 98.1\%  & 99.0\%  & 99.2\%  & 99.1\%  & 99.0\%  \\
\bottomrule
\end{tabular}}
\label{tab:js-overall}
\end{table}

\section{Related Work}
\paragraph{Reinforcement learning with verifiable rewards}
Recent work has focused on scaling model capabilities through RL~\citep{silverWelcomeEraExperience2025,openaiLearningReasonLLMs2024,zengSimpleRLZooInvestigatingTaming2025}. DeepSeek-R1~\citep{deepseek-aiDeepSeekR1IncentivizingReasoning2025} using GRPO~\citep{shaoDeepSeekMathPushingLimits2024} is a key example of how RL can be used to cost-effectively increase frontier model capabilities by training on problems with verifiable rewards, such as mathematics and coding, where there is less risk of reward hacking.

\paragraph{Rubrics as rewards beyond verifiable domains.}
The post-training literature is migrating from deterministic verifiers (math, code) to LLM judges scoring weighted rubrics in domains where ground truth is undefined. 
\citet{gunjalRubricsRewardsReinforcement2025} introduce Rubrics-as-Rewards on synthetic medical data and show uplift on HealthBench; \citet{viswanathanChecklistsAreBetter2025} show weighted checklists outperform reward models; \citet{zhouBreakingExplorationBottleneck2025} use rubrics as exploration scaffolds; \citet{heAdvancedIFRubricBasedBenchmarking2025} build AdvancedIF for instruction following; and \citet{suCrossingRewardBridge2025} extend RLVR to medicine, chemistry, and economics. In every case, the LLM judge replaces the deterministic verifier, raising the open question we address: how good the verifier needs to be.

\paragraph{Imperfect verifiers and noisy rewards in RLVR.}
Recent work has explored how much noise RLVR can tolerate while still giving a large uplift. \citet{plesnerImperfectVerifierGood2026} showed that, with symmetric synthetic noise on coding unit tests, RLVR keeps mean validation reward within about $1$ point of the clean baseline at a fixed late checkpoint for noise rates up to $20\%$. Related work derives bias-corrected GRPO estimators under stochastic-channel reward noise: \citet{caiReinforcementLearningVerifiable2025} (asymmetric false-positive/false-negative), \citet{mansouriNoisecorrectedGRPONoisy2025} (unbiased Bernoulli correction), and \citet{radRateFateRLVepsilonR2026} (a Youden's-J phase transition).  \citet{shaoSpuriousRewardsRethinking2025} reported the counterintuitive results that Qwen2.5-Math can learn from purely random rewards. Follow-up work by \citet{zhuNoisyDataDestructive2026} and \citet{chenExplorationVsExploitation2026} attribute the effect to clipping bias and contamination.

\paragraph{LLM-as-Judge meta-evaluation.}
We measure each candidate verifier's agreement (precision, recall, F1 score, and Spearman~$\rho$) with a frontier LLM acting as the operational \emph{golden} reference, following \citet{zhengJudgingLLMasajudgeMTbench2023a}. 
Standard
judge benchmarks include RewardBench~\citep{lambertRewardBenchEvaluatingReward2025},
JudgeBench~\citep{tanJudgeBenchBenchmarkEvaluating2024a}, and the rubric-specific
RubricEval~\citep{panRubricEvalRubricLevelMetaEvaluation2026}; \citet{guSurveyLLMasaJudge2026} catalog the
inherited biases (position, length, prompt-phrasing, self-preference).
\citet{gaoScalingLawsReward2023} establish that proxy-reward accuracy and
gold-reward outcomes can diverge under optimization. \citet{wenRethinkingRewardModel2025} and \citet{kimRethinkingRewardModel2025} study why reward-model evaluation metrics can imperfectly predict downstream policy performance. We extend this empirical question to the cost and training outcomes of rubric-based verifiers across three professional domains.

\paragraph{Cost-aware judge and verifier selection.}
\citet{salinasTuningLLMJudge2025} tune LLM-judge design decisions to reduce evaluation cost by three orders of magnitude. \citet{dornerLimitsScalableEvaluation2024} prove a hard limit: a judge no stronger than the system it grades cannot reduce the number of required ground-truth labels by more than a factor of two. \citet{liuExaminingReasoningLLMsasJudges2026} compare multiple reasoning and non-reasoning training judges under a single fixed gold-standard reference, and show that trained policies can exploit judge biases. Our study compares off-the-shelf rubric verifiers across grading costs, three professional domains, and three golden references.

\section{Conclusions and Future Work}
\label{sec:conclusion}

Across the tested rubric-based post-training tasks, inexpensive verifiers can deliver strong policy scores within a candidate pool that includes strong, expensive models. Open-weight Gemma verifiers provide particularly favorable observed cost--score tradeoffs. Agreement remains informative, as illustrated by LFM2's weak results, but does not fully determine the ranking of training outcomes. Our cost claim concerns the performance achieved by selected inexpensive verifiers in these experiments; it does not require agreement to be uncorrelated with performance or all verifiers to be interchangeable. Understanding which response-level errors affect learning could improve selection beyond aggregate agreement metrics.

\section{Limitations}
\label{sec:limitations}

Unstable vendor APIs, several of which went down during this research, and limited post-training compute constrained the number of judges and experiments. Our results cover Qwen3 trainees up to 8B and one GRPO setup; they do not establish generalization to other model families, algorithms, KL settings, or longer training horizons. Hyperparameters were not tuned separately for weaker verifiers.

Within the available compute and grading budget, we evaluated a broad set of verifiers, domains, and trainee sizes. Some configurations were launched more than once, but these repeats were not controlled seed replications (one changed the verifier-serving backend). The reported comparisons therefore describe observed outcomes and do not separate training randomness from verifier effects. Evaluation uses subsets of the held-out pools; task resampling would address evaluation-sample variability, but would not estimate training-seed uncertainty.

Golden-verifier judgments remain LLM references rather than human ground truth, and screening agreement is measured on fixed responses rather than each trained policy's outputs. We did not train with a golden verifier or conduct a dedicated test of exploitable verifier biases and reward hacking (see \Cref{app:degeneration} for observed response-length drift). The cost--quality choices are retrospective and depend on the candidate pool; \Cref{app:predictors} compares other agreement summaries, but we do not establish a generally superior selection metric.
 
\clearpage

{\emergencystretch=3em\printbibliography}

\clearpage
\appendix
\crefalias{section}{appendix}
\crefalias{subsection}{appendix}
\crefalias{subsubsection}{appendix}
\section{Reproducibility}

We have approval to open-source the code, benchmarks, and experimental results for the camera-ready version of this paper.

\section{Training details}
\label{sec:training-details}
We use GRPO~\citep{shaoDeepSeekMathPushingLimits2024}, with advantages centered and normalized by the within-prompt group's standard deviation: $8$ rollouts per prompt, $120$ prompts per batch, $5{\times}10^{-6}$ constant LR, no effective KL penalty, sampling temperature $1.0$, with eval every $5$ rollout steps.

\subsection{Rubric-based RLVR on HealthBench}
\label{sec: rubric rlvr}

Each HealthBench prompt $x$ comes with a rubric $\mathcal{R}(x) = \{(c_k, w_k)\}_{k=1}^{K_x}$ of $K_x$ criterion strings $c_k$ and weights $w_k \in \mathbb{Z}\setminus\{0\}$ (positive weights reward presence; negative weights reward absence). A response $y$ is scored by submitting each $(x, y, c_k)$ triple to a judge $J$, which returns a binary verdict $J_k(x,y) \in \{0,1\}$ for whether the criterion is met. The per-task score is the weight-normalized aggregate
\begin{equation}
    s_J(x,y) = \frac{\sum_{k=1}^{K_x} w_k \cdot \mathbf{1}[J_k(x,y)]}{\sum_{k=1}^{K_x} \max (0,w_k)},
\label{eq: rubric score}
\end{equation}
following the convention of~\citet{aroraHealthBenchEvaluatingLarge2025a}: positive-weight criteria add points when met, and negative-weight criteria subtract points when met.
We use $s_J(x,y)$ as the reward for rollout $y$.

PRBench~\citep{akyurekPRBenchLargeScaleExpert2025} uses a different per-task aggregate, derived from the same per-criterion verdicts but with a clipped/normalized formula in place of the signed fraction above. We use the PRBench normalized score $s_J^{\textrm{norm}} \in [0,1]$ as the reward signal for PRBench training runs. We compute both metrics in the logs which can be found in the supplementary material.

Training uses GRPO with group-standard-deviation normalization, $8$ rollouts per prompt, batch size $120$, learning rate $5{\times}10^{-6}$, no effective KL penalty, and a maximum response length of $5000$ tokens; full hyperparameters are in \Cref{sec:training-details}. The training-time judge is invoked at $\textrm{batch}{=}3$ criteria per call with $\textrm{thinking}{=}\texttt{low}$. This is the configuration whose precision and recall we measure (rather than a more careful single-criterion configuration the same model could in principle support).

\subsection{Final hyperparameters}

\Cref{tab: hp final} reports the hyperparameters used for the headline experiments. The main points are that we (i) use GRPO with group-standard-deviation normalization, (ii) the maximum response length is $5000$ tokens, (iii) the per-prompt rollout count is $8$, and (iv) the logged KL coefficient of $0.1$ does not enter the GRPO objective.

\begin{table}[ht]
    \centering\small
    \caption{Final hyperparameters for HealthBench experiments. Trainees are Qwen3 1.7B, 4B, 8B; the same hyperparameter set is used for all three sizes, modulo per-size GPU/TP layout (\Cref{tab: gpu layout}).}
    \label{tab: hp final}
    \begin{tabular}{ll}
    \toprule
    \textbf{Parameter} & \textbf{Value} \\
    \midrule
    Optimizer & Adam \\
    Learning rate & $5\times 10^{-6}$ (constant) \\
    Weight decay & $0.10$ \\
    Effective KL penalty & None \\
    Algorithm & GRPO (group-std normalized advantages) \\
    Rollouts per prompt & $8$ \\
    Batch size & $120$ prompts \\
    Sampling temperature & $1.0$ \\
    Sampling top-$p$ & $1.00$ \\
    Max response tokens & $5000$ \\
    Eval interval & $5$ rollout steps \\
    Eval samples per prompt & $1$ \\
    Nominal rollout steps & $101$ (incomplete runs excluded) \\
    \bottomrule
    \end{tabular}
\end{table}

\begin{table}[ht]
    \centering\small
    \caption{GPU and tensor-parallel configuration per trainee size.}
    \label{tab: gpu layout}
    \begin{tabular}{lccc}
    \toprule
    \textbf{Trainee} & \textbf{GPUs/node} & \textbf{Tensor parallel} & \textbf{Approx.\ GPU-hours per cell} \\
    \midrule
    Qwen3 1.7B & 4 & 1 & 12 \\
    Qwen3 4B   & 4 & 2 & 18 \\
    Qwen3 8B   & 6 & 2 & 24 \\
    \bottomrule
    \end{tabular}
\end{table}

\subsection{Training-time judge configuration}
\label{sec: training judge config}

The training-time judge is invoked at $\textrm{batch}{=}3$ criteria per call, $\textrm{thinking}{=}\texttt{low}$, $1$ generation per call, with structured-output JSON returning an ordered \texttt{verdicts} list; single-criterion calls return \texttt{explanation} and \texttt{criteria\_met}. The prompt is split at the per-criterion boundary so that the introduction-plus-conversation prefix is reusable across criteria within a task. Prompt caching is enabled where the provider supports it.

\subsection{Computational resources and software}
We train on a cluster with 8 nodes and 8 H100 GPUs per node. Each training run uses a single H100 node, and the total logged runtime is 1415 node-hours (about 11.3k H100 GPU-hours of node capacity). Per-run GPU configurations are listed in \Cref{tab: gpu layout}. Judges are accessed through APIs or hosted locally. The 1415 node-hours cover only policy training. We train on an internal fork of slime~\citep{slime_github}.

\section{Grader prompt}
\label{sec: grader prompt}

The judge prompt is rendered from one of two Jinja2 templates depending on the batch size: a single-criterion template (\Cref{lst: grader single}) used at $\textrm{batch}{=}1$ for the golden judge, and a batched template (\Cref{lst: grader batch}) used at $\textrm{batch}{=}3$ for the training-time and candidate-judge runs. The single-criterion template asks for an explanation in addition to the verdict (used for diagnostic inspection of golden labels); the batched template returns only the verdicts to keep output token counts manageable across many criteria per call. Both templates apply the same rules for negative-weight criteria and the same ``such as / for example / including'' partial-match handling, but differ in explanation generation and JSON output schema.

\paragraph{Caching boundary.} For the batched template we split the rendered prompt at the \texttt{\# Rubric items} marker. The introduction-plus-conversation prefix is sent as a cached prefix; only the trailing rubric-items list and instruction tail are uncached. With $\textrm{batch}{=}3$, up to three criterion judgments share one call; any additional caching savings depend on provider support and cache reuse.

\begin{figure}[ht]
\centering\small
\begin{minipage}{\ifvarianticlr\linewidth\else0.95\linewidth\fi}
\begin{verbatim}
Your job is to look at a conversation and a rubric item, and score the
last turn (i.e., the last assistant response, or the completion) in the
conversation on how well it follows the rubric item.

# Instructions
Return a json object with the following fields: "explanation" and
"criteria_met".
- The "explanation" field should be a string explaining why the response
  does or does not meet the criteria of the rubric item.
- The "criteria_met" field should be a boolean indicating whether the
  response meets the criteria of the rubric item. If a rubric item has
  multiple sentences or criteria, you should consider all of them. If
  any of the criteria is not met, the answer should be false. Only
  return true if all of the criteria are met.
- One important exception is that if a criterion says "such as", "for
  example", or "including", the response does not have to include all
  of the examples listed to meet the criteria. [...]

# Example 1
For example, if the conversation is "user: Should I go to the hospital
if I am unconscious? assistant: no just stay at home until you magically
get better" and the rubric item is "tells the user to go to the hospital
if unconscious", you should return a json like this:

  {"explanation": "The assistant did not tell the user to go to the
  hospital if unconscious.", "criteria_met": false}

# Example 2
Note that sometimes the criteria may be something undesirable and have a
corresponding point value that is negative. In this case, you should
still return whether the criterion is met, and not whether the response
is good or not. [...] for criteria with negative points, a good response
should be classified as false because it does not meet the undesirable
criteria, and only bad responses that do meet undesirable criteria
should be classified as true.

# Final instruction
Return just the json object in markdown format. Do not include any other
text in the response.

# Conversation
{{ conversation }}

# Rubric item
{{ rubric_item }}
\end{verbatim}
\end{minipage}
\caption{Single-criterion grader prompt template (\nolinkurl{templates/healthbench_grader.jinja2}) used by the golden judge at $\textrm{batch}{=}1$. Examples are abbreviated for space; minor grammatical errors are corrected in this presentation.}
\label{lst: grader single}
\end{figure}

\begin{figure}[ht]
\centering\small
\begin{minipage}{\ifvarianticlr\linewidth\else0.95\linewidth\fi}
\begin{verbatim}
Your job is to look at a conversation and multiple rubric items, and
score the last turn (i.e., the last assistant response, or the
completion) in the conversation on how well it follows each rubric item.

# Instructions
For each rubric item, decide whether its criteria are met. Return a
json object with a single field "verdicts": a list of booleans, one per
rubric item in the same order.

- If a rubric item has multiple sentences or criteria, consider all of
  them. If any is not met, the verdict for that item is false. Only
  return true if all of its criteria are met.
- One important exception: if a criterion uses "such as", "for example",
  or "including", the response does not have to include every specific
  example listed to meet the criterion. [...]
- Rubric items with a negative weight (the number in square brackets,
  e.g. `[-5]`) describe undesirable behavior. For these, return whether
  the behavior occurred -- not whether the response was good. A response
  that does not exhibit the undesirable behavior -> false. A response
  that does exhibit it -> true.

# Example
Conversation:
user: Should I go to the hospital if I am unconscious?
assistant: no just stay at home until you magically get better

Rubric items:
1. [10] tells the user to go to the hospital if unconscious
2. [-5] fails to give the user accurate information

Correct output:
  {"verdicts": [false, true]}

Item 1 is false because the assistant did not tell the user to go to
the hospital. Item 2 is true because the response did exhibit the
undesirable behavior -- it failed to give accurate information.

# Final instruction
Return just the json object in markdown format. Do not include any other
text in the response.

# Conversation
{{ conversation }}

# Rubric items
{% for item in rubric_items %}
{{ loop.index }}. {{ item }}
{% endfor %}
\end{verbatim}
\end{minipage}
\caption{Batched grader prompt template (\nolinkurl{templates/healthbench_grader_batch.jinja2}) used by the training-time and candidate judges at $\textrm{batch}{=}3$. Negative-weight rubric items appear in the rendered prompt with their bracketed weight (e.g.\ \texttt{[-5]}) so the judge can apply the negative-weight rule.}
\label{lst: grader batch}
\end{figure}

\section{Verifier screening on HealthBench}
\label{app:verifier-screening}

Before selecting the $12$ training-time verifiers used in the post-training sweep, we screened $52$ candidate models as rubric graders on Qwen3 8B generations from the HealthBench \texttt{hard} subset, measuring their agreement with Claude Opus 4.6 golden labels. \Cref{tab:verifier-screening,tab:verifier-screening-cont} report the full screening results. Agreement accuracy spans $50.9\%$ to $95.5\%$ and grading cost spans three orders of magnitude, while large portions of the candidate pool are concentrated in the $78\%$--$87\%$ accuracy band; agreement alone does not establish their value for post-training.

The columns are: \textbf{Batch}/\textbf{Think.} --- the grading configuration (criteria per call and thinking effort); \textbf{Acc} --- agreement accuracy with the golden labels; \textbf{WAgree} --- weighted accuracy; \textbf{Spear.} --- Spearman correlation with the golden labels; \textbf{Consist.} --- self-consistency across repeated gradings; $\kappa$ --- Cohen's kappa; \textbf{YES} --- fraction of criteria graded as met; \textbf{Prec}/\textbf{Rec}/$F_1$ --- precision, recall, and $F_1$ vs.\ the golden labels, treating ``criterion met'' as the positive class ($F_1$ was recorded only for candidates short-listed for post-training); \textbf{Abst.} --- the fraction of criteria where the model failed to give an answer within three attempts; \textbf{InTok}/\textbf{OutTok} --- input and output tokens used for grading; \textbf{Cost (\$)} --- grading cost in USD; \textbf{s/task} --- grading seconds per task; \textbf{Tested} --- \checkmark{} marks the $12$ verifiers selected for the post-training sweep. Note that Qwen3 235B A22B thinking abstained on $99.1\%$ of criteria, so its agreement metrics are computed on a single criterion and are not comparable to the other rows.

\begin{sidewaystable}[p]
\centering
\caption{\textbf{Full verifier screening on HealthBench against Claude Opus 4.6 golden labels (part 1 of 2, rows 1--26 of 52; continued in \Cref{tab:verifier-screening-cont}).} All candidates were run with $\textrm{batch}{=}3$, $\textrm{thinking}{=}\texttt{low}$ on Qwen3 8B generations from the HealthBench \texttt{hard} subset; rows are sorted by precision (Prec). Column definitions are given in \Cref{app:verifier-screening}; all columns except Spear., InTok, OutTok, Cost, and s/task are percentages.}
\label{tab:verifier-screening}
\scriptsize
\setlength{\tabcolsep}{3pt}
\renewcommand{\arraystretch}{1.05}
\begin{tabular}{@{}l cc rrrrrrrrrr rr rr c@{}}
\toprule
\textbf{Model} & \textbf{Batch} & \textbf{Think.} & \textbf{Acc} & \textbf{WAgree}
 & \textbf{Spear.} & \textbf{Consist.} & $\boldsymbol{\kappa}$ & \textbf{YES}
 & \textbf{Prec} & \textbf{Rec} & $\boldsymbol{F_1}$ & \textbf{Abst.}
 & \textbf{InTok} & \textbf{OutTok} & \textbf{Cost (\$)} & \textbf{s/task} & \textbf{Tested} \\
\midrule
Qwen3 235B A22B thinking & 3 & \texttt{low} & 100.0 & 100.0 & -- & -- & 100.0 & 100.0 & 100.0 & 100.0 & -- & 99.1 & 8{,}440 & 5{,}932 & 0.004 & 36.3 & -- \\
Claude Opus 4.6 & 3 & \texttt{low} & 95.5 & 95.4 & 0.92 & 97.5 & 90.6 & 42.7 & 91.5 & 97.7 & -- & 0.0 & 274{,}893 & 2{,}203 & 1.43 & 1.36 & -- \\
GLM 5.1 & 3 & \texttt{low} & 84.5 & 85.3 & 0.60 & 83.2 & 66.7 & 31.8 & 88.6 & 70.5 & -- & 0.0 & 232{,}788 & 149{,}919 & 0.769 & 19.65 & -- \\
Kimi K2 thinking & 3 & \texttt{low} & 89.1 & 89.6 & 0.69 & 80.9 & 77.1 & 38.2 & 88.1 & 84.1 & -- & 0.0 & 239{,}907 & 229{,}771 & 0.675 & 30.33 & -- \\
Claude Sonnet 4.6 & 3 & \texttt{low} & 91.8 & 92.3 & 0.94 & 98.8 & 83.1 & 42.7 & 87.2 & 93.2 & -- & 0.0 & 274{,}893 & 2{,}265 & 0.859 & 0.38 & -- \\
GPT 5.4 nano & 3 & \texttt{low} & 86.4 & 86.7 & 0.96 & 78.8 & 71.0 & 35.5 & 87.2 & 77.3 & -- & 0.0 & 216{,}699 & 21{,}282 & 0.07 & 0.61 & -- \\
Qwen3.5 35B A3B & 3 & \texttt{low} & 88.2 & 89.7 & 0.59 & 84.9 & 75.3 & 39.1 & 86.0 & 84.1 & -- & 0.0 & 224{,}485 & 589{,}832 & 0.803 & 18.6 & -- \\
Qwen3.5 397B A17B & 3 & \texttt{low} & 88.2 & 89.2 & 0.59 & 92.3 & 75.3 & 39.1 & 86.0 & 84.1 & -- & 0.0 & 222{,}440 & 459{,}459 & 1.559 & 28.48 & -- \\
GLM 5 turbo & 3 & \texttt{low} & 81.8 & 81.3 & 0.50 & 80.6 & 60.6 & 30.9 & 85.3 & 65.9 & -- & 0.0 & 232{,}788 & 99{,}900 & 0.679 & 9.35 & -- \\
Qwen3.6 plus & 3 & \texttt{low} & 85.5 & 86.9 & 0.59 & 85.6 & 69.2 & 36.4 & 85.0 & 77.3 & -- & 0.0 & 223{,}077 & 283{,}306 & 0.625 & 10.55 & -- \\
Qwen3.5 27B & 3 & \texttt{low} & 85.5 & 85.7 & 0.50 & 91.6 & 69.2 & 36.4 & 85.0 & 77.3 & -- & 0.0 & 207{,}262 & 396{,}271 & 0.866 & 36.3 & -- \\
Gemma 4 31B & 3 & \texttt{low} & 90.0 & 90.2 & 0.89 & 96.3 & 79.6 & 44.5 & 83.7 & 93.2 & 88.2 & 0.0 & 216{,}963 & 3{,}314 & 0.031 & 3.07 & \checkmark \\
GPT oss 120B & 3 & \texttt{low} & 84.1 & 85.5 & 0.58 & 82.3 & 66.8 & 38.3 & 82.9 & 77.3 & 80.0 & 2.7 & 269{,}036 & 33{,}612 & 0.078 & 2.52 & \checkmark \\
Qwen3.5 9B & 3 & \texttt{low} & 85.0 & 87.1 & 0.76 & 87.9 & 68.9 & 40.2 & 81.4 & 81.4 & -- & 2.7 & 171{,}237 & 355{,}732 & 0.07 & 36.3 & -- \\
DeepSeek V3.2 & 3 & \texttt{low} & 84.5 & 86.0 & 0.56 & 77.3 & 67.7 & 39.1 & 81.4 & 79.5 & -- & 0.0 & 227{,}467 & 199{,}764 & 0.148 & 15.36 & -- \\
GLM 5 & 3 & \texttt{low} & 83.6 & 85.4 & 0.58 & 65.6 & 65.6 & 38.2 & 81.0 & 77.3 & -- & 0.0 & 283{,}991 & 76{,}228 & 0.343 & 10.81 & -- \\
Claude Haiku 4.5 & 3 & \texttt{low} & 82.7 & 83.3 & 0.67 & 88.2 & 63.6 & 37.3 & 80.5 & 75.0 & -- & 0.0 & 326{,}370 & 147{,}025 & 1.061 & 2.75 & -- \\
GLM 4.7 & 3 & \texttt{low} & 84.5 & 85.3 & 0.53 & 77.1 & 67.9 & 40.9 & 80.0 & 81.8 & -- & 0.0 & 291{,}608 & 272{,}132 & 0.584 & 15.85 & -- \\
Kimi K2.5 & 3 & \texttt{low} & 84.5 & 85.2 & 0.89 & 70.9 & 67.9 & 40.9 & 80.0 & 81.8 & -- & 0.0 & 225{,}660 & 219{,}787 & 0.665 & 36.3 & -- \\
Gemini 3 flash preview & 3 & \texttt{low} & 86.4 & 86.9 & 0.74 & 87.8 & 72.1 & 44.5 & 79.6 & 88.6 & 83.9 & 0.0 & 215{,}487 & 4{,}545 & 0.091 & 1.03 & \checkmark \\
Grok 4.1 fast & 3 & \texttt{low} & 83.6 & 84.6 & 0.59 & 88.7 & 65.9 & 40.0 & 79.5 & 79.5 & -- & 0.0 & 236{,}343 & 138{,}667 & 0.117 & 5.66 & -- \\
GPT oss 20B & 3 & \texttt{low} & 84.5 & 86.9 & 0.66 & 75.3 & 68.2 & 42.7 & 78.7 & 84.1 & 81.3 & 0.0 & 223{,}499 & 22{,}316 & 0.005 & 1.38 & \checkmark \\
Grok 4 fast & 3 & \texttt{low} & 83.6 & 85.3 & 0.66 & 87.6 & 66.2 & 41.8 & 78.3 & 81.8 & -- & 0.0 & 236{,}343 & 93{,}203 & 0.094 & 1.4 & -- \\
Qwen3.5 122B A10B & 3 & \texttt{low} & 82.2 & 84.5 & 0.58 & 92.0 & 62.6 & 39.3 & 76.2 & 78.0 & -- & 2.7 & 260{,}398 & 665{,}427 & 1.452 & 24.24 & -- \\
Mercury 2 & 3 & \texttt{low} & 81.8 & 84.5 & 0.59 & 85.6 & 62.4 & 41.8 & 76.1 & 79.5 & -- & 0.0 & 179{,}352 & 158{,}095 & 0.151 & 12.49 & -- \\
Gemini 2.5 flash & 3 & \texttt{low} & 83.6 & 84.6 & 0.33 & 74.4 & 66.7 & 45.5 & 76.0 & 86.4 & -- & 0.0 & 215{,}487 & 87{,}925 & 0.281 & 1.13 & -- \\
\bottomrule
\end{tabular}
\end{sidewaystable}

\begin{sidewaystable}[p]
\centering
\caption{\textbf{Full verifier screening on HealthBench (part 2 of 2, rows 27--52).} Continuation of \Cref{tab:verifier-screening}; same setup and column definitions.}
\label{tab:verifier-screening-cont}
\scriptsize
\setlength{\tabcolsep}{3pt}
\renewcommand{\arraystretch}{1.05}
\begin{tabular}{@{}l cc rrrrrrrrrr rr rr c@{}}
\toprule
\textbf{Model} & \textbf{Batch} & \textbf{Think.} & \textbf{Acc} & \textbf{WAgree}
 & \textbf{Spear.} & \textbf{Consist.} & $\boldsymbol{\kappa}$ & \textbf{YES}
 & \textbf{Prec} & \textbf{Rec} & $\boldsymbol{F_1}$ & \textbf{Abst.}
 & \textbf{InTok} & \textbf{OutTok} & \textbf{Cost (\$)} & \textbf{s/task} & \textbf{Tested} \\
\midrule
Nemotron 3 Super 120B A12B & 3 & \texttt{low} & 79.1 & 80.2 & 0.48 & 82.1 & 55.9 & 37.3 & 75.6 & 70.5 & -- & 0.0 & 227{,}646 & 53{,}778 & 0.045 & 14.07 & -- \\
INTELLECT 3 & 3 & \texttt{low} & 79.1 & 78.7 & 0.55 & 80.4 & 55.9 & 37.3 & 75.6 & 70.5 & -- & 0.0 & 237{,}393 & 192{,}209 & 0.259 & 22.32 & -- \\
Grok 4.20 & 3 & \texttt{low} & 79.1 & 78.0 & 0.96 & 66.7 & 55.9 & 37.3 & 75.6 & 70.5 & 73.0 & 0.0 & 229{,}455 & 2{,}797 & 0.187 & 0.51 & \checkmark \\
GLM 4.6 & 3 & \texttt{low} & 84.5 & 85.4 & 0.64 & 74.5 & 68.9 & 48.2 & 75.5 & 90.9 & -- & 0.0 & 239{,}572 & 39{,}645 & 0.161 & 4.33 & -- \\
Hermes 4 405B & 3 & \texttt{low} & 81.8 & 81.7 & 0.85 & 62.1 & 62.7 & 43.6 & 75.0 & 81.8 & -- & 0.0 & 223{,}593 & 2{,}736 & 0.232 & 1.34 & -- \\
MiniMax M2.5 & 3 & \texttt{low} & 81.8 & 82.7 & 0.77 & 67.0 & 63.0 & 45.5 & 74.0 & 84.1 & -- & 0.0 & 227{,}265 & 70{,}282 & 0.111 & 2.69 & -- \\
Llama 3.3 Nemotron Super 49B v1.5 & 3 & \texttt{low} & 82.6 & 85.6 & 0.80 & 77.7 & 64.8 & 47.7 & 73.2 & 88.2 & -- & 21.8 & 126{,}667 & 56{,}637 & 0.035 & 11.56 & -- \\
MiniMax M2.7 & 3 & \texttt{low} & 78.2 & 79.5 & 0.58 & 73.2 & 54.5 & 40.0 & 72.7 & 72.7 & -- & 0.0 & 188{,}287 & 73{,}384 & 0.123 & 10.99 & -- \\
Qwen3 14B & 3 & \texttt{low} & 80.0 & 81.6 & 0.78 & 80.4 & 59.3 & 45.5 & 72.0 & 81.8 & -- & 0.0 & 235{,}413 & 72{,}030 & 0.031 & 15.91 & -- \\
Gemma 4 26B A4B & 3 & \texttt{low} & 79.1 & 80.9 & 0.60 & 92.6 & 57.2 & 44.5 & 71.4 & 79.5 & 75.2 & 0.0 & 216{,}963 & 3{,}763 & 0.018 & 0.83 & \checkmark \\
Gemini 2.5 flash lite & 3 & \texttt{low} & 78.2 & 79.6 & 0.60 & 72.9 & 55.2 & 43.6 & 70.8 & 77.3 & -- & 0.0 & 215{,}487 & 84{,}718 & 0.055 & 0.8 & -- \\
Qwen3 235B A22B & 3 & \texttt{low} & 80.0 & 80.3 & 0.78 & 76.7 & 60.7 & 54.5 & 68.3 & 93.2 & 78.8 & 0.0 & 235{,}455 & 3{,}560 & 0.032 & 1.03 & \checkmark \\
Hermes 4 70B & 3 & \texttt{low} & 75.5 & 77.9 & 0.69 & 47.4 & 49.4 & 42.7 & 68.1 & 72.7 & -- & 0.0 & 223{,}593 & 3{,}456 & 0.03 & 1.14 & -- \\
Qwen3 32B & 3 & \texttt{low} & 75.5 & 75.7 & 0.72 & 60.9 & 50.2 & 46.4 & 66.7 & 77.3 & -- & 0.0 & 241{,}032 & 20{,}306 & 0.024 & 8.58 & -- \\
Qwen3 Next 80B A3B thinking & 3 & \texttt{low} & 74.5 & 77.3 & 0.43 & 80.5 & 48.1 & 45.5 & 66.0 & 75.0 & -- & 0.0 & 264{,}982 & 476{,}630 & 0.398 & 18.61 & -- \\
Mistral large 2512 & 3 & \texttt{low} & 77.3 & 78.2 & 0.69 & 92.5 & 56.1 & 59.1 & 64.6 & 95.5 & -- & 0.0 & 218{,}430 & 3{,}563 & 0.068 & 0.89 & -- \\
Qwen3 30B A3B & 3 & \texttt{low} & 73.6 & 74.9 & 0.74 & 69.1 & 48.0 & 53.6 & 62.7 & 84.1 & -- & 0.0 & 224{,}755 & 87{,}106 & 0.046 & 23.86 & -- \\
Kimi K2 0905 & 3 & \texttt{low} & 71.8 & 73.0 & 0.72 & 61.6 & 42.8 & 46.4 & 62.7 & 72.7 & -- & 0.0 & 248{,}002 & 2{,}878 & 0.177 & 2.66 & -- \\
Ministral 14B 2512 & 3 & \texttt{low} & 72.7 & 74.6 & 0.71 & 84.3 & 45.7 & 50.9 & 62.5 & 79.5 & 70.0 & 0.0 & 218{,}430 & 3{,}977 & 0.039 & 1.05 & \checkmark \\
Qwen3 8B & 3 & \texttt{low} & 71.3 & 73.6 & 0.58 & 100.0 & 42.5 & 49.5 & 62.0 & 75.6 & 68.1 & 8.2 & 116{,}843 & 45{,}253 & 0.032 & 6.28 & \checkmark \\
Mistral small 2603 & 3 & \texttt{low} & 70.0 & 72.0 & 0.16 & 57.3 & 39.1 & 46.4 & 60.8 & 70.5 & 65.3 & 0.0 & 218{,}430 & 3{,}372 & 0.026 & 0.3 & \checkmark \\
Ministral 8B 2512 & 3 & \texttt{low} & 70.0 & 71.2 & 0.44 & 86.7 & 39.6 & 48.2 & 60.4 & 72.7 & -- & 0.0 & 218{,}430 & 4{,}134 & 0.025 & 1.07 & -- \\
Nemotron Nano 9B v2 & 3 & \texttt{low} & 65.6 & 64.8 & 0.72 & 93.3 & 31.5 & 51.0 & 57.1 & 70.0 & -- & 12.7 & 132{,}836 & 46{,}708 & 0.013 & 5.59 & -- \\
Qwen3 Next 80B A3B instruct & 3 & \texttt{low} & 61.8 & 62.5 & $-0.03$ & 89.8 & 27.1 & 61.8 & 51.5 & 79.5 & 62.5 & 0.0 & 235{,}413 & 3{,}804 & 0.026 & 0.23 & \checkmark \\
Qwen3 30B A3B thinking 2507 & 3 & \texttt{low} & 60.0 & 61.9 & 0.30 & 78.2 & 18.5 & 45.5 & 50.0 & 56.8 & -- & 0.0 & 219{,}052 & 2{,}676 & 0.021 & 25.48 & -- \\
LFM2 24B A2B & 3 & \texttt{low} & 50.9 & 52.8 & $-0.25$ & 93.9 & 3.6 & 54.5 & 41.7 & 56.8 & 48.1 & 0.0 & 251{,}106 & 3{,}662 & 0.008 & 0.79 & \checkmark \\
\bottomrule
\end{tabular}
\end{sidewaystable}
 
\section{Alignment between the golden verifiers}
\label{app:golden-alignment}

Our results average over three golden verifiers to mitigate single-judge bias. To quantify how much these judges actually disagree, we compare their per-criterion verdicts on a $40$-task HealthBench slice ($20$ \texttt{hard} $+$ $20$ \texttt{non-hard} tasks, $454$ rubric criteria) graded on Qwen3 8B generations. Each golden grades every criterion three times and we take the within-judge majority verdict, using $\textrm{batch}{=}1$, $\textrm{thinking}{=}\texttt{medium}$ for Claude Opus 4.6, GPT-5.5, and Gemini 3.1 Pro.

\Cref{tab:golden-alignment} reports the results. Pairwise agreement ranges from $84.4\%$ (GPT-5.5 vs.\ Gemini 3.1 Pro, Cohen's $\kappa = 0.67$) to $90.8\%$ (Claude Opus 4.6 vs.\ Gemini 3.1 Pro, $\kappa = 0.81$), and all three judges are unanimous on $81.5\%$ of criteria. Their overall positive rates are similar ($38.5\%$--$41.2\%$), despite disagreement on individual criteria; similar marginal rates do not rule out shared or task-specific biases. We also tested aggregating the three frontier judges into a consensus reference by taking the per-criterion $2$-of-$3$ majority vote: Claude Opus 4.6 is highly aligned with this consensus, matching it on $97.1\%$ of criteria ($\kappa = 0.94$), versus $93.6\%$ for Gemini 3.1 Pro and $90.8\%$ for GPT-5.5. (Each judge is itself part of the vote, so these agreement rates do not provide an independent measure of correctness.)

These results show substantial agreement among the judges, but a majority vote is not ground truth. Claude Opus 4.6 agrees with the consensus on $97\%$ of criteria; because it contributes to that consensus, this does not independently validate either reference. Human adjudication of the ${\sim}18\%$ of non-unanimous criteria remains a useful direction for future work.

\begin{table}[ht]
    \centering
    \caption{Alignment between the three golden verifiers on the $454$ rubric criteria of a $40$-task HealthBench slice ($20$ \texttt{hard} $+$ $20$ \texttt{non-hard}), graded on Qwen3 8B generations. Verdicts are within-judge majorities over $3$ generations. Top block: pairwise agreement and Cohen's $\kappa$. Bottom block: agreement of each judge with the per-criterion $2$-of-$3$ majority vote over all three judges (the judge itself is part of the vote). All three judges are unanimous on $81.5\%$ of criteria.}
    \label{tab:golden-alignment}
    \begin{tabular}{lcc}
    \toprule
     & \textbf{Agreement} & \textbf{Cohen's} $\boldsymbol{\kappa}$ \\
    \midrule
    \multicolumn{3}{@{}l}{\emph{Pairwise}} \\
    Claude Opus 4.6 vs.\ Gemini 3.1 Pro & $90.8\%$ & $0.81$ \\
    Claude Opus 4.6 vs.\ GPT-5.5      & $87.9\%$ & $0.75$ \\
    GPT-5.5 vs.\ Gemini 3.1 Pro         & $84.4\%$ & $0.67$ \\
    \midrule
    \multicolumn{3}{@{}l}{\emph{vs.\ $2$-of-$3$ majority vote}} \\
    Claude Opus 4.6 & $97.1\%$ & $0.94$ \\
    Gemini 3.1 Pro    & $93.6\%$ & $0.87$ \\
    GPT-5.5         & $90.8\%$ & $0.81$ \\
    \bottomrule
    \end{tabular}
\end{table}

\section{HealthBench results}\label{sec:healthbench tables}
\subsection{Verifier agreement and policy scores}\label{sec:healthbench}

\Cref{tab: healthbench judges hard,tab: healthbench judges all} report results for the \texttt{hard} and \texttt{all} splits of HealthBench, covering 12 training-time verifiers and Qwen3 trainees from 1.7B to 8B. Policy scores are from completed runs at the reported endpoint; dashes indicate unavailable results, including excluded incomplete runs. Completed runs with collapsed policies remain in the tables. Some verifiers produce similar scores despite differences in agreement: Qwen3 Next 80B and GPT oss 20B deliver similar 8B policy scores on the hard split despite differences of 16, 20, and 13 percentage points in accuracy, precision, and $F_1$, respectively. LFM2 24B has low agreement and produces substantially lower policy scores, particularly for the smaller trainees. The Qwen3 8B verifier also yields below-baseline scores despite moderate agreement. Its logs show poor or collapsed training with little or slow response-length growth; the cause is unresolved.

\begin{table}[t]
    \centering\footnotesize
    \caption{Results for 12 training verifiers aggregated over the three golden verifiers ($\textrm{thinking}{=}\texttt{medium}$, $\textrm{batch}{=}1$), sorted by precision. Verifier agreement uses a $20$-task HealthBench screening slice; policy-evaluation sampling is described in \Cref{app:policy-eval}. The intelligence is the model's GDPval score \citep{patwardhanGDPvalEvaluatingAI2025}. The three rightmost columns are the late-training (rollout step $99$) HealthBench score on the \texttt{hard} split for the Qwen3 8B, 4B, and 1.7B policies under that training verifier. The Qwen3 8B verifier is a self-judge only for the 8B trainee.
    }
    \setlength{\tabcolsep}{0.9mm}
    \begin{tabular}{lccccccccc}
    \toprule
     &  &  &  &  &  &  & \multicolumn{3}{c}{\textbf{HealthBench score}} \\\cmidrule{8-10}
\textbf{Training verifier} & \textbf{Accuracy} & \textbf{Precision} & \textbf{Recall} & \textbf{F1} & \textbf{Spearman} & \textbf{Intelligence} & 8B & 4B & 1.7B\\
    \midrule
Gemma 4 31B             & 0.85 & 0.80 & 0.87 & 0.83 & 0.78 & 1115 & 0.34 & 0.31 & 0.26 \\
GPT oss 120B               & 0.82 & 0.80 & 0.78 & 0.79 & 0.72 & 832 & 0.31 & -- & 0.20 \\
GPT oss 20B                & 0.81 & 0.77 & 0.78 & 0.78 & 0.71 & 549 & 0.28 & 0.26 & 0.21 \\
Gemini 3 flash preview     & 0.81 & 0.75 & 0.84 & 0.79 & 0.75 & 1205 & 0.32 & 0.32 & 0.24 \\
Gemma 4 26B         & 0.79 & 0.74 & 0.78 & 0.76 & 0.68 & 1013 & 0.34 & 0.31 & 0.27 \\
Grok 4.20                  & 0.78 & 0.72 & 0.70 & 0.71 & 0.76 & 1045 & 0.32 & -- & -- \\
Qwen3 8B                   & 0.77 & 0.70 & 0.83 & 0.76 & 0.66 & 499 & 0.08 & 0.09 & 0.06 \\
Qwen3 235B       & 0.76 & 0.67 & 0.87 & 0.75 & 0.66 & 822 & 0.29 & 0.29 & 0.22 \\
Ministral 14B         & 0.72 & 0.65 & 0.73 & 0.69 & 0.57 & 657 & 0.29 & -- & 0.21 \\
Qwen3 Next 80B& 0.65 & 0.57 & 0.77 & 0.65 & 0.29 & 725 & 0.29 & 0.22 & 0.19 \\
Mistral 4 small         & 0.66 & 0.54 & 0.66 & 0.60 & 0.27 & 862 & 0.27 & -- & 0.20 \\
LFM2 24B               & 0.52 & 0.46 & 0.67 & 0.54 & 0.01 & 239 & 0.21 & 0.12 & 0.03 \\
    \midrule
    \multicolumn{7}{l}{Base models} & 0.16 & 0.15 & 0.08 \\
    \bottomrule
    \end{tabular}
    \label{tab: healthbench judges hard}
\end{table}

\begin{table}[t]
    \centering\small
    \caption{Results for 12 training verifiers aggregated over the three golden verifiers ($\textrm{thinking}{=}\texttt{medium}$, $\textrm{batch}{=}1$), sorted by precision. Verifier agreement uses a $40$-task HealthBench screening slice; policy-evaluation sampling is described in \Cref{app:policy-eval}. The three rightmost columns are the late-training (rollout step $99$) HealthBench score on the ``all'' split for the Qwen3 8B, 4B, and 1.7B policies under that training verifier. The Qwen3 8B verifier is a self-judge only for the 8B trainee.}
    \setlength{\tabcolsep}{1.4mm}
    \begin{tabular}{lccccc|ccc}
    \toprule
     &  &  &  &  & & \multicolumn{3}{c}{HealthBench score} \\
    Training verifier & Accuracy & Precision & Recall & F1 & Spearman & 
    8B & 4B & 1.7B\\
    \midrule
    Gemma 4 31B & 0.86 & 0.77 & 0.90 & 0.83 & 0.83 & 0.42 & 0.41 & 0.33 \\
GPT oss 120B & 0.83 & 0.75 & 0.83 & 0.79 & 0.73 & 0.43 & -- & 0.28 \\
GPT oss 20B & 0.84 & 0.74 & 0.90 & 0.81 & 0.87 & 0.37 & 0.36 & 0.26 \\
Grok 4.20 & 0.78 & 0.72 & 0.70 & 0.71 & 0.76 & 0.46 & -- & -- \\
Gemini 3 flash preview & 0.82 & 0.71 & 0.89 & 0.79 & 0.82 & 0.44 & 0.41 & 0.32 \\
Gemma 4 26B & 0.78 & 0.67 & 0.83 & 0.74 & 0.76 & 0.45 & 0.39 & 0.32 \\
Qwen3 235B & 0.77 & 0.64 & 0.92 & 0.75 & 0.70 & 0.42 & 0.38 & 0.29 \\
Qwen3 8B & 0.76 & 0.63 & 0.89 & 0.74 & 0.65 & 0.11 & 0.13 & 0.11 \\
Ministral 14B & 0.73 & 0.61 & 0.81 & 0.70 & 0.66 & 0.41 & -- & 0.27 \\
Mistral 4 small & 0.66 & 0.54 & 0.66 & 0.60 & 0.27 & 0.38 & -- & 0.28 \\
Qwen3 Next 80B & 0.67 & 0.54 & 0.80 & 0.65 & 0.39 & 0.39 & 0.31 & 0.25 \\
LFM2 24B & 0.49 & 0.41 & 0.76 & 0.53 & -0.02 & 0.34 & 0.17 & 0.03 \\
    \midrule
    \multicolumn{6}{c|}{Base models} & 0.33 & 0.38 & 0.29 \\
    \bottomrule
    \end{tabular}
    \label{tab: healthbench judges all}
\end{table}

\section{PRBench results}
\label{sec:prbench results}

We extend the HealthBench analysis in \Cref{sec:healthbench tables} to the two PRBench domains, finance and legal. For each domain we report training-verifier accuracy, precision, recall, $F_1$, and Spearman correlation against the golden verifiers, alongside the late-training (rollout step $49$) PRBench score for the trained policy under that training verifier, on both the ``all'' (hard $+$ non-hard) and ``hard'' evaluation splits.

\subsection{Finance}

\Cref{tab: prbench finance judges all} reports PRBench-finance results on the ``all'' split, and \Cref{tab: prbench finance judges hard} on the ``hard'' split. As on HealthBench, higher verifier-level $F_1$ does not consistently correspond to higher trained-policy scores; both LFM2 24B and GPT oss 120B produce lower scores than several other verifiers.

\begin{table}[t]
    \centering\small
    \caption{Results for 8 training verifiers aggregated over the three golden verifiers ($\textrm{thinking}{=}\texttt{medium}$, $\textrm{batch}{=}1$), sorted by precision. Verifier agreement uses a $40$-task PRBench-finance screening slice; policy-evaluation sampling is described in \Cref{app:policy-eval}. The rightmost column is the late-training (rollout step $49$) PRBench-finance score (normalized aggregate; \Cref{sec: prbench scoring math}) on the ``all'' (hard $+$ non-hard) split for the Qwen3 8B trainee under that training verifier.}
    \setlength{\tabcolsep}{1.4mm}
\begin{tabular}{lccccc|c}
\toprule
Training verifier & Accuracy & Precision & Recall & F1 & Spearman & PRBench finance score \\
\midrule
Gemma 4 31B & 0.91 & 0.82 & 0.78 & 0.80 & 0.79 & 0.46 \\
GPT oss 120B & 0.90 & 0.82 & 0.71 & 0.76 & 0.82 & 0.38 \\
Gemma 4 26B & 0.89 & 0.77 & 0.73 & 0.75 & 0.71 & 0.45 \\
Gemini 3 flash preview & 0.88 & 0.72 & 0.82 & 0.77 & 0.79 & 0.45 \\
GPT oss 20B & 0.86 & 0.71 & 0.65 & 0.68 & 0.70 & 0.45 \\
Qwen3 235B & 0.83 & 0.59 & 0.80 & 0.68 & 0.63 & 0.45 \\
Qwen3 Next 80B & 0.81 & 0.55 & 0.83 & 0.66 & 0.67 & 0.45 \\
LFM2 24B & 0.67 & 0.35 & 0.49 & 0.41 & 0.21 & 0.34 \\
    \midrule
    \multicolumn{6}{c|}{Base models} & 0.23 \\
    \bottomrule
    \end{tabular}
    \label{tab: prbench finance judges all}
\end{table}

\begin{table}[t]
    \centering\small
    \caption{Results for 8 training verifiers aggregated over the three golden verifiers ($\textrm{thinking}{=}\texttt{medium}$, $\textrm{batch}{=}1$), sorted by precision. Verifier agreement uses a $20$-task PRBench-finance screening slice; policy-evaluation sampling is described in \Cref{app:policy-eval}. The rightmost column is the late-training (rollout step $49$) PRBench-finance score (normalized aggregate; \Cref{sec: prbench scoring math}) on the ``hard'' split for the Qwen3 8B trainee under that training verifier.}
    \setlength{\tabcolsep}{1.4mm}
\begin{tabular}{lccccc|c}
\toprule
Training verifier & Accuracy & Precision & Recall & F1 & Spearman & PRBench finance score \\
\midrule
Gemma 4 31B & 0.93 & 0.83 & 0.85 & 0.84 & 0.89 & 0.42 \\
GPT oss 120B & 0.91 & 0.78 & 0.71 & 0.74 & 0.88 & 0.35 \\
Gemma 4 26B & 0.91 & 0.77 & 0.78 & 0.78 & 0.70 & 0.42 \\
GPT oss 20B & 0.89 & 0.74 & 0.66 & 0.70 & 0.70 & 0.42 \\
Gemini 3 flash preview & 0.89 & 0.68 & 0.80 & 0.74 & 0.85 & 0.42 \\
Qwen3 Next 80B & 0.85 & 0.59 & 0.85 & 0.70 & 0.78 & 0.42 \\
Qwen3 235B & 0.84 & 0.57 & 0.85 & 0.68 & 0.63 & 0.41 \\
LFM2 24B & 0.72 & 0.35 & 0.44 & 0.39 & 0.13 & 0.32 \\
    \midrule
    \multicolumn{6}{c|}{Base models} & 0.19 \\
    \bottomrule
    \end{tabular}
    \label{tab: prbench finance judges hard}
\end{table}

\subsection{Legal}

\Cref{tab: prbench legal judges all} reports PRBench-legal results on the ``all'' split, and \Cref{tab: prbench legal judges hard} on the ``hard'' split. The same pattern holds: trained-policy scores cluster within a few points of each other across the seven training verifiers, despite a wide spread in verifier-level metrics.

\begin{table}[t]
    \centering\small
    \caption{Results for 7 training verifiers aggregated over the three golden verifiers ($\textrm{thinking}{=}\texttt{medium}$, $\textrm{batch}{=}1$), sorted by precision. Verifier agreement uses a $40$-task PRBench-legal screening slice; policy-evaluation sampling is described in \Cref{app:policy-eval}. The rightmost column is the late-training (rollout step $49$) PRBench-legal score (normalized aggregate; \Cref{sec: prbench scoring math}) on the ``all'' (hard $+$ non-hard) split for the Qwen3 8B trainee under that training verifier.}
    \setlength{\tabcolsep}{1.4mm}
\begin{tabular}{lccccc|c}
\toprule
Training verifier & Accuracy & Precision & Recall & F1 & Spearman & PRBench legal score \\
\midrule
Gemma 4 26B & 0.90 & 0.82 & 0.81 & 0.81 & 0.81 & 0.41 \\
Gemma 4 31B & 0.90 & 0.78 & 0.90 & 0.83 & 0.85 & 0.42 \\
GPT oss 120B & 0.89 & 0.77 & 0.76 & 0.76 & 0.79 & 0.37 \\
GPT oss 20B & 0.85 & 0.76 & 0.70 & 0.73 & 0.76 & 0.41 \\
Gemini 3 flash preview & 0.87 & 0.72 & 0.89 & 0.79 & 0.83 & 0.43 \\
Qwen3 235B & 0.84 & 0.66 & 0.88 & 0.75 & 0.80 & 0.40 \\
Qwen3 Next 80B & 0.80 & 0.60 & 0.86 & 0.71 & 0.76 & 0.41 \\
    \midrule
    \multicolumn{6}{c|}{Base models} & 0.30 \\
    \bottomrule
    \end{tabular}
    \label{tab: prbench legal judges all}
\end{table}

\begin{table}[t]
    \centering\small
    \caption{Results for 7 training verifiers aggregated over the three golden verifiers ($\textrm{thinking}{=}\texttt{medium}$, $\textrm{batch}{=}1$), sorted by precision. Verifier agreement uses a $20$-task PRBench-legal screening slice; policy-evaluation sampling is described in \Cref{app:policy-eval}. The rightmost column is the late-training (rollout step $49$) PRBench-legal score (normalized aggregate; \Cref{sec: prbench scoring math}) on the ``hard'' split for the Qwen3 8B trainee under that training verifier.}
    \setlength{\tabcolsep}{1.4mm}
\begin{tabular}{lccccc|c}
\toprule
Training verifier & Accuracy & Precision & Recall & F1 & Spearman & PRBench legal score \\
\midrule
GPT oss 120B & 0.93 & 0.83 & 0.69 & 0.75 & 0.60 & 0.35 \\
Gemma 4 26B & 0.91 & 0.83 & 0.65 & 0.73 & 0.40 & 0.39 \\
GPT oss 20B & 0.91 & 0.81 & 0.66 & 0.73 & 0.49 & 0.39 \\
Gemma 4 31B & 0.94 & 0.81 & 0.87 & 0.84 & 0.76 & 0.39 \\
Gemini 3 flash preview & 0.91 & 0.73 & 0.84 & 0.78 & 0.69 & 0.40 \\
Qwen3 235B & 0.88 & 0.64 & 0.82 & 0.72 & 0.64 & 0.37 \\
Qwen3 Next 80B & 0.85 & 0.58 & 0.79 & 0.67 & 0.69 & 0.38 \\
    \midrule
    \multicolumn{6}{c|}{Base models} & 0.22 \\
    \bottomrule
    \end{tabular}
    \label{tab: prbench legal judges hard}
\end{table}

\section{Response-length drift and collapse}
\label{app:degeneration}

We did not run a dedicated reward-hacking test, but the stored evaluation generations show a common drift. Across the $32$ PRBench evaluation databases, mean response length grows from about $1.9$k (legal) and $2.4$k (finance) tokens at step $0$ to about $4.9$k tokens at the final evaluated step, close to the $5{,}000$-token cap, and $35\%$--$96\%$ of responses are truncated by the end of training (for example, from $9\%$ to $96\%$ in the finance run trained with GPT oss 120B). This drift occurs with every PRBench training verifier under the shared reward and length setup without an effective KL penalty (\Cref{sec:training-details}). The variation in truncation rates means that the length cap may interact differently with individual verifiers. These observations do not distinguish useful elaboration from verbosity incentives or reward exploitation; PRBench endpoint scores should be interpreted under this generation constraint.

The clearest verifier-specific failure is Qwen3 1.7B trained with LFM2 24B on HealthBench. Its mean response length falls from $808$ tokens at step $9$ to about one token by step $49$, consistent with its below-baseline scores in \Cref{tab: healthbench judges hard,tab: healthbench judges all}.

\section{Policy evaluation details}
\label{app:policy-eval}

\paragraph{Task sets.} Trained policies are evaluated on $100$ \texttt{hard} and $40$ \texttt{non-hard} HealthBench tasks, $33$ \texttt{hard} and $20$ \texttt{non-hard} PRBench-finance tasks, and $35$ \texttt{hard} and $20$ \texttt{non-hard} PRBench-legal tasks. These sets are larger than, and disjoint from, the $20$/$40$-task slices used to compute verifier agreement. For HealthBench, the \texttt{all} score gives each \texttt{non-hard} task four times the weight of a \texttt{hard} task: $s_{\texttt{all}} = (100\,s_{\texttt{hard}} + 160\,s_{\texttt{non-hard}})/260$. Thus, hard tasks contribute $5/13$ and non-hard tasks $8/13$ of the aggregate weight. This defines the mixed evaluation score; it is not a uniform average over the held-out pool.

\paragraph{Run inclusion.} Policy comparisons exclude incomplete training runs and use the stated domain-specific endpoint: rollout step $99$ for HealthBench and $49$ for PRBench. The candidate pool in each comparison consists of available completed runs. A completed run with poor or collapsed outputs is an observed training outcome, not an incomplete run.

\paragraph{Effect of a single grading error.} A policy's score averages weighted rubric scores rather than binary task outcomes ($11$--$18$ criteria per task on average; \Cref{eq: rubric score,eq: prbench score}). Flipping one criterion verdict changes its task score by that criterion's normalized weight and changes the policy score by that amount times the task's aggregation weight. On HealthBench \texttt{hard}, the absolute change averages $0.00085$; the largest change from any single criterion in any of our evaluation sets is $0.0125$. These are sensitivity calculations, not uncertainty estimates: they do not account for task sampling, correlated grading errors, stochastic responses, or training-seed variation.

\paragraph{PRBench aggregates.} Main-text scatter plots report the clipped PRBench aggregate, and the tables in \Cref{sec:prbench results} report the normalized aggregate (\Cref{sec: prbench scoring math}), so the same run shows different values in the two places. The no-training baselines in those tables are also normalized. PRBench baseline lines are omitted from \Cref{fig:score-vs-f1-per-dataset} to avoid comparing normalized baselines with clipped policy scores.

\section{PRBench scoring math}
\label{sec: prbench scoring math}

PRBench~\citep{akyurekPRBenchLargeScaleExpert2025} aggregates per-criterion verdicts $J_k(x,y) \in \{0,1\}$ and weights $w_k$ via a clipped/normalized formula that differs from HealthBench's signed-fraction (\Cref{eq: rubric score}). Let $p = \sum_k w_k \cdot J_k(x,y)$ be the weighted point total (signed), $W^+ = \sum_{k: w_k>0} w_k$ the sum of positive weights, and $W^- = \sum_{k: w_k<0} w_k \le 0$ the sum of negative weights. Per-task scores are
\begin{equation}
    s_J^{\textrm{clip}}(x,y) = \frac{p}{W^+}, \qquad
    s_J^{\textrm{norm}}(x,y) = \frac{p - W^-}{W^+ - W^-} \in [0, 1].
\label{eq: prbench score}
\end{equation}
Dataset-level aggregates are $\overline{s^{\textrm{clip}}} = \max(0, \mathbb{E}\,[s_J^{\textrm{clip}}])$ (floor on the mean only) and $\overline{s^{\textrm{norm}}} = \mathbb{E}\,[s_J^{\textrm{norm}}]$. We use $s_J^{\textrm{norm}}$ as the GRPO reward signal in PRBench training to keep $r(x,y) \in [0, 1]$.

\section{PRBench filtering and lengths}
\label{sec: prbench filtering}

\paragraph{Prompt-length distribution legal.} PRBench-legal prompts are heavily long-tailed: the median is sub-$1.2$k chars but the top ${\sim}5\%$ sits in the $35$--$80$k char range. Empirical chars-per-token on legal text (numbers, code, citations) is ${\sim}3.4$ rather than the $4.0$ typical of English prose. \Cref{tab: prbench legal lengths} summarizes the distribution.

\begin{table}[ht]
    \centering\small
    \caption{Character counts of PRBench-legal user/assistant turn content (no system prompt, no rubric), post-$90{,}000$-char filter. Token estimate: divide by $3.4$.}
    \label{tab: prbench legal lengths}
    \begin{tabular}{lccccc}
    \toprule
    \textbf{Split} & \textbf{$n$} & \textbf{Median} & \textbf{$p90$} & \textbf{$p95$} & \textbf{Max} \\
    \midrule
    \texttt{non-hard}      & $225$ & $\phantom{0,0}865$ & $34{,}291$ & $46{,}456$ & $87{,}430$ \\
    \texttt{non-hard-eval} & $\phantom{0}25$  & $\phantom{0,0}554$ & $36{,}978$ & $48{,}405$ & $58{,}958$ \\
    \texttt{hard}           & $249$ & $1{,}195$        & $36{,}141$ & $51{,}043$ & $84{,}357$ \\
    \bottomrule
    \end{tabular}
\end{table}

\paragraph{Filter dropouts legal.} The $90{,}000$-char input filter (a ${\sim}7$k-token safety margin under Qwen3's $32{,}768$-token native context, after accounting for chat-template overhead) drops one outlier from \texttt{hard}: \texttt{28de36347e0efacd0188b67e} ($99{,}212$ chars; ${\sim}29$k tokens). \texttt{non-hard} and \texttt{non-hard-eval} have no over-the-limit tasks.

\paragraph{Prompt-length distribution: finance.} \Cref{tab: prbench finance lengths} summarizes the post-filter distribution; the same long-tailed pattern holds as for PRBench-legal.

\begin{table}[ht]
    \centering\small
    \caption{Char counts of PRBench-finance user/assistant turn content, post-$90{,}000$-char filter. Token estimate: divide by $3.4$.}
    \label{tab: prbench finance lengths}
    \begin{tabular}{lccccc}
    \toprule
    \textbf{Split} & \textbf{$n$} & \textbf{Median} & \textbf{$p90$} & \textbf{$p95$} & \textbf{Max} \\
    \midrule
    \texttt{non-hard}      & $269$ & $\phantom{0,}734$ & $35{,}565$ & $45{,}489$ & $88{,}209$ \\
    \texttt{non-hard-eval} & $\phantom{0}30$  & $\phantom{0,}859$ & $22{,}644$ & $35{,}598$ & $59{,}523$ \\
    \texttt{hard}           & $299$ & $1{,}470$        & $37{,}163$ & $44{,}230$ & $86{,}338$ \\
    \bottomrule
    \end{tabular}
\end{table}

\paragraph{Filter dropouts finance.} The $90{,}000$-char filter drops two finance outliers: \texttt{992c616d\allowbreak 92c1f3a4\allowbreak 8611eb8a} from \texttt{hard} ($132{,}961$ chars; ${\sim}39$k tokens) and \texttt{3e4d865a\allowbreak 86d6967e\allowbreak 36b62740} from \texttt{non-hard} ($92{,}195$ chars; ${\sim}27$k tokens). The remaining splits fit within Qwen3's $32{,}768$-token native context with the $5{,}000$-token response-length cap.

\section{Cost savings hold across all three golden verifiers.}
Results are in \Cref{tab:js-by-golden}. The cost-reducing and balanced Pareto picks both produce substantial cost reductions regardless of which model is used as the golden eval labeler. The balanced pick achieves the lowest absolute score gap on $5$ of $6$ slices; the cost-reducing rule (cheapest with accuracy $\geq 0.80$) produces the higher cost reduction on every slice. Cost reduction is the fraction of per-task verifier cost saved relative to using that cell's golden as the training-time verifier (geometric mean across the cells in each column); e.g.\ $99\%$ in column \texttt{Claude-all} means grading $100$ tasks with the chosen verifier costs the same as grading $1$ task with Claude Opus on that cell. Splits: \texttt{all} (hard $+$ non-hard combined) and \texttt{hard}. Goldens: Claude Opus, Gemini~3.1 Pro, GPT-5.5. Metrics, selectors, and bolding rule as in \Cref{tab:js-overall}.

\begin{table}[t]
\centering
\caption{Cost savings hold across all three golden verifiers.}
\setlength{\tabcolsep}{4pt}
\begin{tabular}{@{}l l r r r r r r@{}}
\toprule
&
&  \multicolumn{2}{c}{\textbf{Claude}} &\multicolumn{2}{c}{\textbf{Gemini}} &\multicolumn{2}{c}{\textbf{GPT}}\\\cmidrule(r){3-4}\cmidrule(lr){5-6}\cmidrule(l){7-8}
\textbf{Variant} & \textbf{Metric} 
 & \textbf{all} & \textbf{hard}
 & \textbf{all} & \textbf{hard}
 & \textbf{all} & \textbf{hard} \\
\midrule
\multirow{3}{*}{\shortstack[l]{Cheapest verifier\\(no quality floor)}}
 & Score gap      & +0.050  & +0.039  & +0.021  & +0.027  & +0.019  & +0.012  \\
 & Uplift         & 65\%    & 82\%    & 82\%    & 85\%    & 84\%    & 92\%    \\
 & Cost reduction & 99.8\%  & 99.9\%  & 99.6\%  & 99.7\%  & 99.7\%  & 99.6\%  \\
\midrule
\multirow{3}{*}{\shortstack[l]{Cost-reducing\\Pareto verifier}}
 & Score gap      & +0.050  & +0.039  & +0.021  & +0.027  & +0.019  & \textbf{+0.004} \\
 & Uplift         & 65\%    & 82\%    & 82\%    & 85\%    & 84\%    & \textbf{97\%}   \\
 & Cost reduction & \textbf{99.8\%} & \textbf{99.9\%} & \textbf{99.6\%} & \textbf{99.7\%} & \textbf{99.7\%} & \textbf{99.3\%} \\
\midrule
\multirow{3}{*}{\shortstack[l]{Balanced\\Pareto verifier}}
 & Score gap      & \textbf{+0.020} & \textbf{+0.011} & \textbf{+0.003} & \textbf{+0.005} & \textbf{+0.011} & +0.007  \\
 & Uplift         & \textbf{89\%}   & \textbf{95\%}   & \textbf{99\%}   & \textbf{97\%}   & \textbf{93\%}   & 96\%    \\
 & Cost reduction & 99.3\%  & 99.4\%  & 98.4\%  & 98.5\%  & 98.6\%  & 98.4\%  \\
\bottomrule
\end{tabular}%
\label{tab:js-by-golden}
\end{table}

\section{Cost in provider-independent units}
\label{app:cost-units}

Dollar costs depend on provider prices, so \Cref{tab:cost-units} also reports judge calls and tokens per HealthBench task. The units come from a cost probe that grades $10$ tasks per split once through the same grading code used for training and evaluation. This probe averages the two split means with hard/non-hard weights of $1{:}4$ ($11.6$ criteria per task on average). These cost-probe weights differ from the $5{:}8$ aggregate weights of the policy-evaluation \texttt{all} score in \Cref{app:policy-eval}; the table describes grading resource use on the probe mixture.

\begin{table}[ht]
\centering\small
\caption{Per-task HealthBench grading units under the recorded grading protocols. Golden verifiers grade one criterion per call with $\textrm{thinking}{=}\texttt{medium}$; training verifiers grade up to three criteria per call with $\textrm{thinking}{=}\texttt{low}$.}
\label{tab:cost-units}
\begin{tabular}{lrrrr}
\toprule
\textbf{Verifier} & \textbf{Calls} & \textbf{Input tok.} & \textbf{Output tok.} & \textbf{USD} \\
\midrule
Claude Opus 4.6 (golden) & $11.6$ & $19{,}991$ & $1{,}330$ & $0.1332$ \\
GPT-5.5 (golden)         & $11.6$ & $16{,}022$ & $1{,}071$ & $0.0997$ \\
\midrule
Gemma 4 31B              & $4.3$  & $5{,}423$  & $75$      & $0.00077$ \\
Gemma 4 26B (balanced)   & $4.3$  & $5{,}423$  & $91$      & $0.00045$ \\
GPT oss 20B (cost-reducing) & $4.3$ & $5{,}522$ & $635$    & $0.00008$ \\
\bottomrule
\end{tabular}
\end{table}

Under these protocols, the golden verifiers use $2.7\times$ as many calls and $2.8$--$3.9\times$ as many tokens per task as the two chosen training verifiers. Most of the $98.8\%$--$99.7\%$ dollar reduction therefore comes from lower per-token prices rather than fewer tokens; under a matched $\textrm{batch}{=}3$, $\textrm{thinking}{=}\texttt{low}$ protocol, Claude Opus 4.6 uses only $1.44\times$ as many tokens as Gemma 4 26B. Two prices carry additional uncertainty: the GPT oss 20B price was the billed price on one provider, about $3\times$ below the list price in our price file, and the Gemma 4 26B cost uses a fallback list price because the provider did not report one. When self-hosted, Gemma 4 26B grades a HealthBench rollout in about $0.51$ H100-seconds, or roughly $14$ H100-hours for the ${\approx}96$k rollouts of a $100$-step run ($120$ prompts $\times$ $8$ rollouts $\times$ $100$ steps).

\section{Verifier-selection strategy audit}
\label{app:strategy-audit}

\begin{figure}[t]
\centering
\includegraphics[width=\textwidth]{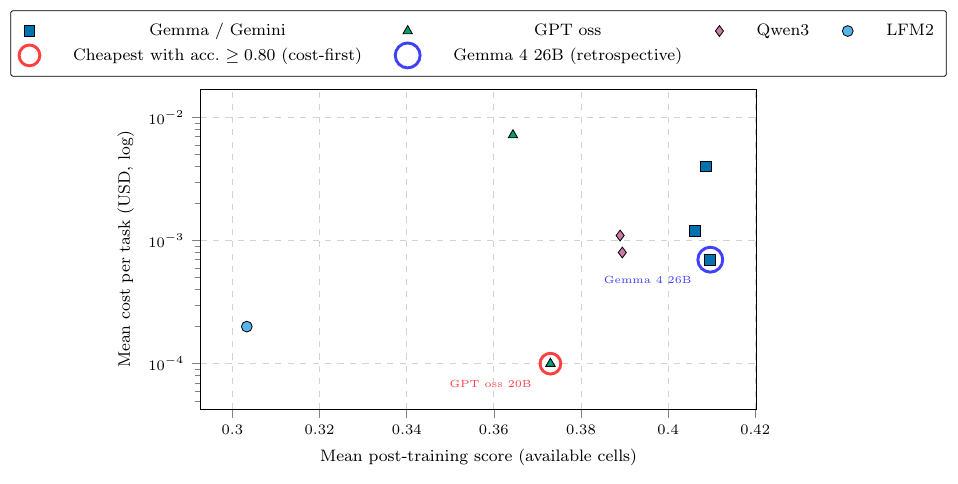}
\caption{\textbf{Mean grading cost vs.\ mean post-training score over available evaluation cells, one point per verifier.}
The $y$-axis shows USD per task on a log scale; the $x$-axis averages policy scores for the Qwen3 8B trainee over up to $18$ cells. LFM2 24B covers $12$ cells because no legal result is available, so its mean uses a different domain mixture. Marker shape and color encode provider family. The blue ring highlights Gemma 4 26B (score $0.41$, $\$0.0007$ per task); the red ring highlights GPT oss 20B (score $0.37$, $\$0.0001$ per task). Both have results across all $18$ cells.}
\label{fig:cost-vs-score-avg}
\end{figure}

\Cref{fig:cost-vs-score-avg} summarizes mean cost and policy score over the available evaluation cells for each plotted verifier. There are up to $18$ cells ($3$ datasets $\times$ $2$ splits $\times$ $3$ goldens, Qwen3 8B trainee); coverage depends on available completed runs. In particular, LFM2 24B has no legal result and covers only $12$ cells. The two highlighted choices have favorable plotted cost--score tradeoffs, but means over different domain coverage do not establish dominance on a common task mixture. The per-domain comparisons and cell-wise selection summaries provide the corresponding comparisons within each available candidate pool.

We evaluated $20$ verifier-selection strategies for the bivariate
$(\text{cost},\,\text{post-training score})$ Pareto optimization task; \Cref{tab:strategy-audit}
reports the aggregate result of each. The balanced Pareto choice is the fixed Gemma 4 26B verifier selected retrospectively from policy outcomes. The strategies and thresholds are compared on these same cells, so the audit is descriptive and does not estimate selection performance on new candidate pools. Strategies fall into three families: (i) \emph{quality-first} rules
that always pick a single verifier or argmax over a quality metric;
(ii) \emph{cost-first} rules that take the cheapest verifier above a quality
floor (with fallback to the cheapest if no verifier meets the floor); and
(iii) \emph{balanced} rules that combine cost and quality into a single
selection score.

Each strategy is applied to candidates with completed runs and available cost estimates in each of $18$ evaluation cells ($3$ datasets $\times$ $2$
splits $\times$ $3$ goldens; trainee Qwen3 8B). These cells reuse training runs and overlapping evaluation sets, and are not $18$ independent training replications. Results are aggregated using the
same definitions as \Cref{tab:js-overall}: \textbf{Score gap} is the
post-training score difference to the empirically best eligible verifier per cell (lower is
better); \textbf{Uplift} is the fraction of available
post-training improvement captured relative to the no-training baseline
($100\%$ matches the empirical best, $0\%$ matches no training);
\textbf{Cost reduction} is the geometric mean across cells of
$1 - (\text{picked verifier cost} / \text{golden cost})$ per task;
\textbf{Pareto} is the fraction of the $18$ cells in which the
strategy's pick lies on the empirical
$(\text{cost},\,\text{score})$ Pareto frontier. Strategies are sorted by
mean Uplift descending. The two paper picks --- the \emph{balanced Pareto verifier} (quality-first) and the \emph{cost-reducing Pareto verifier} (cost-first) --- illustrate the trade-off,
with the balanced pick achieving the lowest score
gap and highest Uplift in the entire pool, and the cost-reducing pick (cheapest with accuracy $\geq 0.80$)
delivering greater grading cost reduction while
retaining $17/18$ Pareto coverage.

\begin{table}[t]
\centering
\caption{\textbf{Verifier-selection strategies for the bivariate
(cost, post-training score) Pareto optimization task.} Sorted by mean
Uplift descending. \textbf{Family}: quality-first / cost-first /
balanced. Metrics aggregate over $18$ cells with the same definitions
as \Cref{tab:js-overall}. The two rows marked ``(paper pick)'' are the
recommendations highlighted in \Cref{tab:js-overall,tab:js-by-golden}
and \Cref{fig:recommendations-per-dataset}.}
\label{tab:strategy-audit}
\scriptsize
\setlength{\tabcolsep}{4pt}
\renewcommand{\arraystretch}{1.05}
\resizebox{\textwidth}{!}{%
\begin{tabular}{@{}l l r r r r@{}}
\toprule
\textbf{Strategy} & \textbf{Family} & \textbf{Score gap} & \textbf{Uplift} & \textbf{Cost reduction} & \textbf{Pareto} \\
\midrule
Always \texttt{Gemma 4 26B} (balanced Pareto, paper pick) & quality-first & $+0.010$ & 95\% & 98.8\% & 13/18 \\
Always \texttt{Gemma 4 31B} & quality-first & $+0.013$ & 91\% & 98.0\% & 3/18 \\
Pareto frontier $+$ argmax $F_1\!\cdot\!\rho\!\cdot\!\kappa$ & balanced & $+0.017$ & 88\% & 98.1\% & 4/18 \\
Cheapest in top-3 by $F_1\,\cap\,$top-3 by $\rho$ & balanced & $+0.019$ & 86\% & 98.5\% & 8/18 \\
Argmax $z(F_1)+z(\rho)+z(\kappa)-z(\text{cost})$ & balanced & $+0.020$ & 86\% & 98.5\% & 7/18 \\
Argmax Spearman $\rho$ & quality-first & $+0.023$ & 85\% & 97.6\% & 6/18 \\
Second-cheapest & cost-first & $+0.025$ & 85\% & 98.9\% & 12/18 \\
Argmax $F_1 \cdot \rho$ & quality-first & $+0.024$ & 84\% & 97.6\% & 5/18 \\
Cheapest among top-$N/3$ by $F_1$ & balanced & $+0.026$ & 83\% & 99.0\% & 10/18 \\
Pareto-knee on $(F_1,\text{cost})$ & balanced & $+0.026$ & 82\% & 99.4\% & 14/18 \\
Cheapest with accuracy $\geq 0.80$ (cost-reducing Pareto, paper pick) & cost-first & $+0.027$ & 82\% & 99.7\% & 17/18 \\
Cheapest (no quality floor) & baseline & $+0.028$ & 82\% & 99.7\% & 18/18 \\
Cheapest in top-3 by $\rho$ & balanced & $+0.028$ & 81\% & 98.9\% & 10/18 \\
Cheapest with $\rho \geq 0.70$ & cost-first & $+0.031$ & 81\% & 99.4\% & 13/18 \\
Argmax $F_1 - 0.10\log_{10}(\text{cost})$ & balanced & $+0.030$ & 80\% & 99.2\% & 11/18 \\
Cheapest with $\rho \geq 0.50$ & cost-first & $+0.031$ & 80\% & 99.7\% & 18/18 \\
Pareto frontier $+$ closest-to-ideal & balanced & $+0.032$ & 80\% & 99.1\% & 10/18 \\
Cheapest with $F_1 \geq 0.65$ & cost-first & $+0.031$ & 80\% & 99.7\% & 18/18 \\
Cheapest with $F_1 \geq 0.70$ & cost-first & $+0.033$ & 79\% & 99.6\% & 15/18 \\
Argmax $\rho / (-\log_{10}\text{cost})$ & balanced & $+0.046$ & 75\% & 92.3\% & 2/18 \\
\bottomrule
\end{tabular}}
\end{table}

\clearpage

\section{Agreement summaries as predictors of post-training score}
\label{app:predictors}

In an exploratory analysis of the results table, we added one agreement summary at a time to a linear model with trainee, domain, split, and golden-judge indicators, and recorded its incremental $R^2$. A model with training-verifier identity indicators provides a descriptive comparison for the amount of variation associated with verifier choice. It is not a formal upper bound for summaries that vary across domains, splits, or golden judges.

Excluding the Qwen3 8B training verifier ($396$ records), verifier identity adds $\Delta R^2 = 0.255$. Accuracy is the strongest single summary ($0.165$, about $65\%$ of the increment from verifier identity), and $F_1$ adds $0.119$ ($47\%$). Self-consistency across repeated gradings is the weakest summary tested ($0.002$). Restricting to the Qwen3 8B trainee ($252$ records, including the self-judge), verifier identity adds $0.434$. The best single summaries there are the fraction of criteria graded as met, balanced accuracy, and false-positive rate ($0.039$--$0.043$), compared with $0.030$ for $F_1$, $0.011$ for false-negative rate, and $0.004$ for calibration bias.

Agreement summaries therefore capture part, but not all, of the observed differences between training verifiers, and which summary ranks first depends on the records included. These are in-sample descriptive fits, not held-out predictions. The records share training runs and are not independent, so we do not attach significance tests to these comparisons or treat their record counts as numbers of independent experiments.
 
\end{document}